\RequirePackage{iftex}
\ifPDFTeX
  \pdfoutput=1
\fi
\documentclass[11pt]{article}

\PassOptionsToPackage{table}{xcolor}
\let\latexaddcontentsline\addcontentsline
\usepackage[final]{acl}
\let\addcontentsline\latexaddcontentsline
\usepackage{times}
\usepackage{latexsym}
\usepackage[T1]{fontenc}
\usepackage[utf8]{inputenc}
\usepackage{microtype}
\usepackage{enumitem}
\usepackage{tocloft}
\usepackage{etoc}
\usepackage{graphicx}
\usepackage{subcaption}
\usepackage{booktabs}
\usepackage{wrapfig}
\usepackage{multirow}
\usepackage{amsmath}
\usepackage{amssymb}
\usepackage{mathtools}
\usepackage{algorithm}
\usepackage{algorithmic}

\usepackage{amsthm}
\usepackage{pifont}
\usepackage{tikz}
\usetikzlibrary{positioning, shapes.misc, arrows.meta}
\usepackage[capitalize,noabbrev]{cleveref}
\usepackage{listings}
\usepackage[most]{tcolorbox}
\usepackage[textsize=tiny]{todonotes}

\definecolor{darkgreen}{RGB}{50,100,0}
\definecolor{darkred}{RGB}{200, 0, 0}
\definecolor{lightred}{RGB}{250, 200, 200}
\definecolor{lightblue}{RGB}{230, 240, 255}
\newcommand{\blue}{\cellcolor{lightblue}}

\newcommand\emoji{\raisebox{-8.0pt}{\includegraphics[width=2.0em]{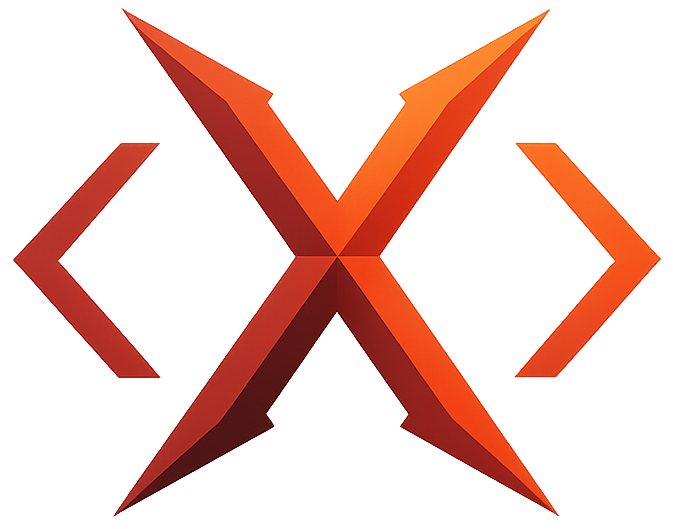}}}
\hypersetup{
    colorlinks,
    citecolor=[rgb]{0.00, 0.45, 0.70},
    linkcolor=[rgb]{0.00, 0.45, 0.70},
    urlcolor=[rgb]{0.80, 0.47, 0.74}
}

\definecolor{keyblue}{RGB}{97,175,239}
\definecolor{valorange}{RGB}{209,154,102}
\definecolor{numcyan}{RGB}{86,156,214}
\definecolor{kwpurple}{RGB}{198,120,221}

\lstdefinelanguage{json}{
    basicstyle=\ttfamily\scriptsize,
    breaklines=true,
    breakatwhitespace=true,
    showstringspaces=false,
    frame=single,
    tabsize=2,
    rulecolor=\color{darkgreen},
    stringstyle=\color{valorange},
    literate=
     *{true}{{{\color{kwpurple}true}}}{4}
      {false}{{{\color{kwpurple}false}}}{5}
      {null}{{{\color{kwpurple}null}}}{4}
      {0}{{{\color{numcyan}0}}}{1}
      {1}{{{\color{numcyan}1}}}{1}
      {2}{{{\color{numcyan}2}}}{1}
      {3}{{{\color{numcyan}3}}}{1}
      {4}{{{\color{numcyan}4}}}{1}
      {5}{{{\color{numcyan}5}}}{1}
      {6}{{{\color{numcyan}6}}}{1}
      {7}{{{\color{numcyan}7}}}{1}
      {8}{{{\color{numcyan}8}}}{1}
      {9}{{{\color{numcyan}9}}}{1}
}

\newcommand{\appendixlisting}[3][]{%
  \tcbinputlisting{%
    enhanced jigsaw,
    breakable,
    listing only,
    colback=black!2,
    colframe=black!35,
    colbacktitle=black!8,
    coltitle=black,
    boxrule=0.35pt,
    arc=1pt,
    left=1mm,
    right=1mm,
    top=0.8mm,
    bottom=0.8mm,
    before skip=6pt,
    after skip=6pt,
    fonttitle=\bfseries\footnotesize,
    title={#2},
    title after break={#2 (continued)},
    listing file={#3},
    listing options={%
      basicstyle=\ttfamily\tiny,
      language={},
      breaklines=true,
      breakatwhitespace=true,
      columns=fixed,
      keepspaces=true,
      showstringspaces=false,
      #1%
    },
  }%
}

\newtcblisting{appendixjsonlisting}{%
  enhanced jigsaw,
  breakable,
  listing only,
  colback=black!2,
  colframe=black!35,
  boxrule=0.35pt,
  arc=1pt,
  left=0.6mm,
  right=0.6mm,
  top=0.6mm,
  bottom=0.6mm,
  before skip=4pt,
  after skip=4pt,
  listing options={%
    language=json,
    basicstyle=\ttfamily\tiny,
    breaklines=true,
    breakatwhitespace=true,
    columns=fixed,
    keepspaces=true,
    showstringspaces=false,
    frame=none,
    xleftmargin=0pt,
    breakautoindent=true,
    breakindent=0pt,
  },
}

\theoremstyle{plain}

\theoremstyle{definition}

\theoremstyle{remark}

\newcommand{\cmark}{\textcolor{darkgreen}{\ding{51}}}
\newcommand{\xmark}{\textcolor{darkred}{\ding{55}}}
\newcommand{\method}{X-Coder}

\title{\texorpdfstring{\emoji-Coder}{X-Coder}: Advancing Competitive Programming with Synthetic Tasks, Solutions, and Tests}

\author{
{\normalfont\normalsize
\textbf{Jie Wu}$^{1,*}$ \quad
\textbf{Haoling Li}$^{1,*}$ \quad
\textbf{Xin Zhang}$^{2,*,\dagger\ddagger}$ \quad
\textbf{Jiani Guo}$^{1}$ \quad
\textbf{Jane Luo}$^{2}$ \quad
\textbf{Xuewei Yang}$^{1}$} \\
{\normalfont\normalsize
\textbf{Steven Liu}$^{2}$ \quad
\textbf{Yangyu Huang}$^{2}$ \quad
\textbf{Ruihang Chu}$^{1,\ddagger}$ \quad
\textbf{Scarlett Li}$^{2}$ \quad
\textbf{Yujiu Yang}$^{1}$} \\
{\normalfont\normalsize $^{1}$Tsinghua University \quad $^{2}$Microsoft} \\
{\normalfont\normalsize $^{*}$Equal Contribution \quad $^{\dagger}$Project Lead \quad $^{\ddagger}$Corresponding Author} \\
{\normalfont\normalsize\texttt{\{wujie24,li-hl23,yang.yujiu\}@mails.tsinghua.edu.cn}}
}

\begin{document}
\maketitle
\begingroup
\renewcommand{\thefootnote}{}
\footnotetext{Work done during the internships of Jie Wu, Haoling Li, Jiani Guo, Jane Luo, and Steven Liu at Microsoft Research.}
\endgroup

\begin{abstract}
Competitive programming remains challenging for code LLMs. Despite recent progress, many training pipelines still depend on scarce real-world data, raising concerns about scalability and near-duplicate benchmark contamination.
In this paper, we examine whether synthetic training artifacts can support the complete SFT-to-RL cycle for competitive programming: no real-world tasks, solutions, or test cases are directly used for post-training.
To this end, we synthesize tasks, verified solutions, and reliable test cases that serve as reward signals for reinforcement learning.
To improve reward reliability, we introduce a dual-verification strategy that reduces noise in both selected solutions and test outputs. Using this high-quality data, we train the X-Coder model series. X-Coder-14B achieves 67.5\% avg@8 on LiveCodeBench v5 and 63.4\% on v6, outperforming its base model by over 40 points. 
Further analysis provides practical insights into synthetic post-training, highlighting the value of diverse tasks, verified long-CoT supervision, and code-centric reinforcement. Our data and models are released at \href{https://github.com/JieWu02/X-Coder}{\texttt{github.com/JieWu02/X-Coder}}.
\end{abstract}

\etocdepthtag.toc{mtchapter}
\begin{figure}[!h]
    \centering
    \includegraphics[width=1.0\linewidth]{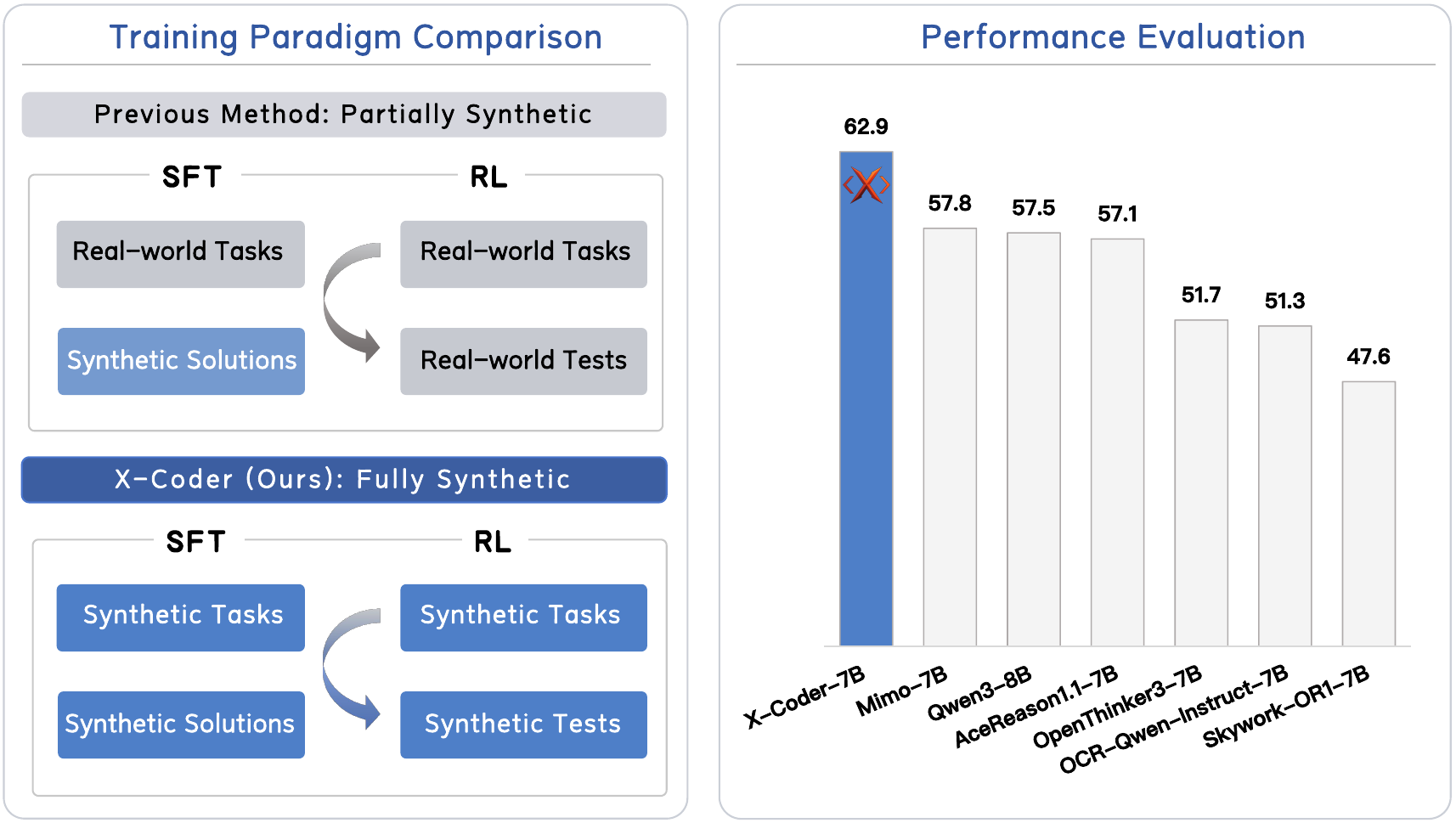}
    \caption{Left: To address persistent data scarcity in competitive programming, we build a scalable alternative by training Code LLMs with synthetic data. Right: Avg@8 results on LiveCodeBench v5. \method~achieves significant performance gains on competitive programming using synthetic data.}
    \label{fig:teaser}
\end{figure}

\section{Introduction}

As Code LLMs advance, they have demonstrated remarkable proficiency in basic programming tasks. Benchmarks such as HumanEval~\citep{chen2021evaluatinglargelanguagemodels,10.5555/3666122.3667065} and MBPP~\citep{austin2021programsynthesislargelanguage} are largely considered solved. However, the frontier of code intelligence has shifted towards competitive programming. This domain is exemplified by platforms like Codeforces and benchmarks like LiveCodeBench~\citep{jain2024livecodebenchholisticcontaminationfree}, requiring deep algorithmic reasoning and complex problem-solving capabilities.

To bridge this gap, recent success in reasoning models has been driven by the utilization of extensive reasoning data, both for supervised fine-tuning (SFT) and reinforcement learning (RL)~\citep{guo2025deepseek,openr1,deepcoder2025}. However, this data-driven scaling faces a critical bottleneck: the scarcity of high-quality training data. The finite pool of real-world competitive programming tasks is insufficient to sustain the scaling laws of reasoning models. Furthermore, existing collections~\citep{hendrycks2021apps,li2022competition} are heavily exhausted, raising concerns about near-duplicate benchmark contamination and overfitting. This dilemma poses a critical question: Can models reach expert-level reasoning without reusing real-world tasks, solutions, or tests in post-training?

\vspace{10pt}

Answering this question is non-trivial. Generating high-quality synthetic data for competitive programming is significantly harder than for general-purpose programming, facing unique challenges across three dimensions:

(i) Tasks must be both solvable and challenging. Synthesizing competitive-level tasks often yields ill-defined or trivial instances, failing to provide the rigorous complexity needed to train expert models.
(ii) High-quality SFT demands logically sound solutions. Validating these without ground-truth test cases is inherently risky, potentially polluting the training set with incorrect reasoning paths.
(iii) RL necessitates reliable reward signals. Weak or insufficient synthetic tests fail to catch incorrect solutions, providing noisy reward signals that can mislead the reinforcement learning process.

Given these challenges, off-the-shelf synthesis frameworks~\citep{wang2025epicoder,wei2024selfcodealign} fail to generalize to competitive programming and yield suboptimal results.
In this work, we validate the hypothesis that synthetic training artifacts can drive expert-level code reasoning. In our setting, no real-world task, solution, or test case is directly used in SFT or RL. The pipeline nevertheless uses real data indirectly: abstract competition-related features are extracted from 10k TACO snippets without reusing task content, and teacher LLMs may have learned from real-world corpora. We investigate the key factors governing synthetic code reasoning data and establish a high-quality foundation through dual verification.

To validate this setting, we train the X-Coder model series using only synthesized artifacts under an SFT-then-RL paradigm. As shown in Figure~\ref{fig:teaser}, X-Coder achieves significant performance gains on the challenging LiveCodeBench v5 and v6.

Through extensive evaluation, we further distill key insights into what makes synthetic data effective, shedding light on synthetic data scaling, task diversity, verified long-CoT supervision, and the role of code-centric reinforcement learning.

Our contributions are as follows:

(1) We formulate SFT-then-RL from synthetic training artifacts for competitive programming.
We examine whether Code LLMs can be post-trained entirely from synthetic tasks, solutions, and tests while accounting for indirect dependence on seed features and teachers.

(2) We develop the X-Coder model series. X-Coder-14B achieves 67.5\% on LiveCodeBench v5 and 63.4\% on v6, outperforming its base model by over 40 points. X-Coder also generalizes across model families and diverse benchmarks.

(3) Through extensive experiments and ablations, we analyze synthetic data scaling, task diversity, long-CoT supervision, verification reliability, and RL initialization, offering practical guidance for future work on synthetic post-training. We release related models and data.

\section{Related Work}

\noindent \textbf{Data Synthesis for Code Generation.}
Synthetic code data has been studied as a way to alleviate the scarcity of high-quality programming tasks. Existing methods improve code data through instruction tuning and evolution~\citep{yu2024wavecoderwidespreadversatileenhancement,luo2024wizardcoder,xu2024wizardlm,liu2025rstarcoderscalingcompetitivecode}, coder--reviewer interaction~\citep{sun2025codeevointeractiondrivensynthesiscodecentric}, concept composition~\citep{wei2024selfcodealign}, and feature-based problem synthesis~\citep{wang2025epicoder}. Recent work moves beyond instruction--code pairs by constructing verifiable question--solution--test triplets or by using multi-round validation and cross-verification~\citep{xu-etal-2025-kodcode,zhou2026autocode}. These studies demonstrate the value of synthetic code data and executable verification. However, existing pipelines either target general code generation or still directly use real-world tasks in post-training. In contrast, our SFT and RL artifacts contain only synthetic tasks, solutions, and tests, although their construction is indirectly seeded by real-data features and teacher models.

\noindent \textbf{Test Generation and Verification.}
Executable tests are important for code reasoning because they provide correctness checks and training feedback for RL. Prior work has studied LLM-based test generation, feedback refinement, solution-based validation, and reward construction for coder RL~\citep{he2025hardtests,wang-etal-2025-codecontests,cai2026codecontestsopoweringllmsfeedbackdriven,fu2025klearcodetestscalabletestcase,zeng-etal-2025-acecoder}. Other studies evaluate test-suite quality, fault exposure, verifier accuracy, and coverage, including proactive bug discovery through repository-level test generation, or explore grammar-guided, efficiency-oriented, generator-based, and adversarial testing~\citep{yang-etal-2025-llms,ma2026rethinking,cao2026llmsgeneratereliabletest,liu2026testexplorabenchmarkingllmsproactive,ijcai2025p861,peng2025coffe,shi2026codehackerautomatedtestcase,hort2025codehacksdatasetadversarialtests}. These works highlight both the value and the difficulty of reliable executable tests. Our~work shares the motivation of scalable test synthesis, but uses generated tests mainly as RL reward signals in an SFT-then-RL pipeline built from synthetic training artifacts. We use dual verification to reduce noise in synthesized tests.

\noindent \textbf{Post-training for Competitive Programming.}
Existing approaches include supervised fine-tuning on real-world tasks or their variants~\citep{bespoke_stratos,guha2025openthoughtsdatarecipesreasoning,liu2025rstarcoderscalingcompetitivecode}, preference-based code alignment that emphasizes functional differences between correct and incorrect programs~\citep{wu2025teachingmodelsunderstandcode}, reinforcement learning with executable or GRPO-like reward optimization~\citep{shao2024deepseekmath,he2025skywork,deepcoder2025,fu2025areal}, and hybrid SFT-then-RL pipelines that often mix coding and mathematical reasoning data~\citep{liu2025acereasonnemotron11advancingmath,xiaomi2025mimounlockingreasoningpotential,su2025klearreasoneradvancingreasoningcapability}. Despite their strong performance, these methods still rely heavily on finite real-world competitive-programming corpora. Our work instead studies whether synthetic tasks, solutions, and executable tests can sustain the complete SFT-then-RL cycle, where synthetic tests provide reward signals for code-centric RL.

\section{Synthesis of Competition-Level Data}
\label{sec:framework}

\begin{figure*}[!t]
    \centering
    \includegraphics[width=0.9\textwidth]{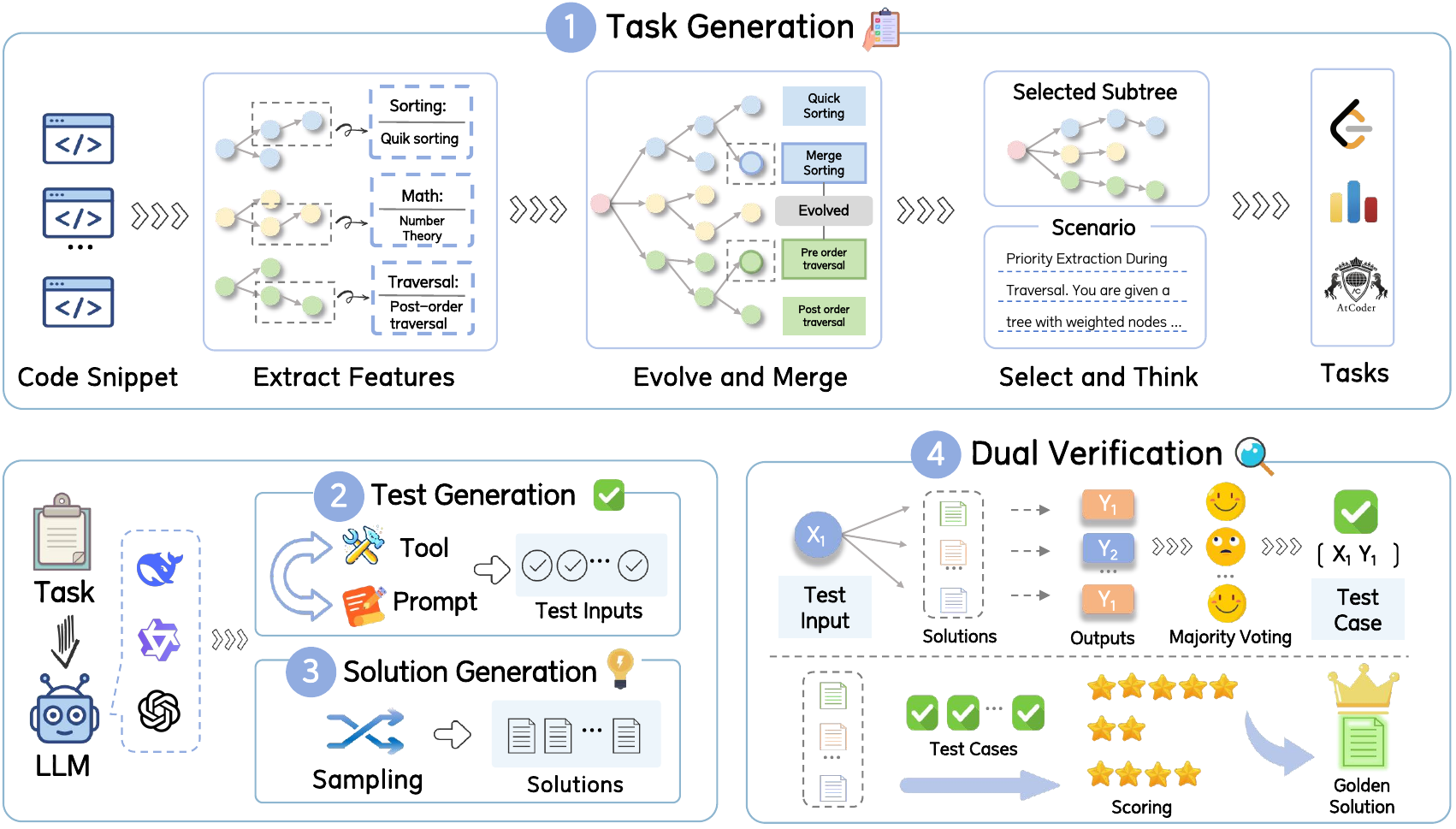}
\caption{\textbf{Overview of the data synthesis framework.} It begins with feature extraction and evolution to construct a domain-specific feature tree. We generate multi-style competitive tasks by formulating scenarios from compatible feature sets. Finally, we synthesize solutions and test inputs, employing a dual-verification strategy to ensure solution correctness and test accuracy, yielding a high-quality data foundation for post-training.}
    \label{fig:framework}
\end{figure*}

We adapt the feature-based synthesis framework~\citep{wang2025epicoder} to the competitive programming domain, establishing a comprehensive methodology to construct tasks that support both SFT and RL stages. As illustrated in Figure~\ref{fig:framework}, this process addresses the unique challenges of synthesis through four key steps:
(i) generating novel and challenging problems (with the capacity for easy scaling in quantity);
(ii) constructing diverse and comprehensive test inputs for each problem (including boundary and stress tests);
(iii) producing high-quality candidate solutions; and
(iv) employing a dual-verification strategy that cross-checks solutions with test cases to yield more accurate test outputs and more reliable solutions.

\noindent \textbf{(i) Task Generation.} While EpiCoder~\citep{wang2025epicoder} introduced feature-based synthesis for general coding tasks, we extend this approach with three key improvements specifically tailored for the complexity of competitive programming.
First, instead of relying on broad definitions of features, we explicitly extract and evolve competition-related features from 10k code snippets in the TACO dataset~\citep{li2023taco} using GPT-4o-0513 (detailed in Appendix~\ref{appendix_sec:Feature Extraction and Evolution}). Second, formulating competitive scenarios from a rich feature tree is non-trivial, as LLMs often oversimplify complex prompts into trivial cases, thereby reducing both diversity and difficulty.
To address this, we adopt a two-stage process that separates feature selection from scenario formulation: first, selecting mutually consistent features for meaningful composition; and second, formulating hint-free tasks that demand genuine reasoning. This two-stage approach significantly outperforms single-step generation (Appendix Table~\ref{single_vs_two_step}). We further incorporate one-shot prompting to improve task understanding and instruction-following. Third, we adapt the synthesis method to support multi-style task generation, covering Codeforces\footnote{\url{https://codeforces.com/}}-style tasks (standard input/output with narrative-rich contexts), LeetCode-style\footnote{\url{https://leetcode.com/}} tasks (starter code with predefined function signatures), and AtCoder\footnote{\url{https://atcoder.jp/}}-style tasks (concise specifications with minimal explanations), thereby enhancing task diversity.
Examples of the task generation process are provided in Appendix~\ref{appendix_sec:Stylized Task Generation for Competitive Programming}, together with difficulty estimates on generated tasks in Appendix~\ref{appendix_sec: Task Difficulty Estimates} and style comparisons in Appendix~\ref{appendix_sec: Ablation on Task Style}.

\noindent \textbf{(ii) Test Input Generation.} 
Obtaining sufficient and accurate test cases is a formidable challenge. Problems from competitive programming platforms often do not provide test cases, or only provide a limited number, due to platform constraints. This results in insufficient quantity, difficulty, and coverage of test cases during RL training.
To address the inherent scarcity of test cases in synthesized data, we investigate two complementary methods for generating the input component of the test case. The \textit{Prompting-based} method instructs the LLM to interpret the problem's input constraints and directly generate multiple test inputs, covering both standard cases and edge-case instances. The \textit{Tool-based} method leverages CYaRon\footnote{\url{https://github.com/luogu-dev/cyaron}}, a dedicated test case generation library, enabling the LLM to construct test inputs by invoking functions documented within the library after understanding the problem.
For each task, we generate a set of $n$ test case inputs $[x_1, x_2, \ldots, x_n]$.
Test input generation is detailed in Appendix~\ref{appendix_sec: test case generation}, with a comparison in Appendix~\ref{appendix_sec: test case comparison}.

\noindent \textbf{(iii) Candidate Solutions Generation.}
For each task, we generate multiple candidate solutions using advanced open-source reasoning LLMs, obtaining $m$ answers $[A^1, A^2, \ldots, A^m]$.
We verify that each candidate solution includes a complete chain-of-thought reasoning process and a syntactically valid Python code implementation, ensuring the absence of syntax errors through static analysis based on Abstract Syntax Tree (AST) parsing. Filtering details are in Appendix~\ref{appendix_sec: Validation on Solution}.

\noindent \textbf{(iv) Dual-Verification of Solutions and Test Cases.} We adopt a dual-verification strategy. Step 1 of this strategy extends the principle of self-consistency~\citep{wang2022self} by applying majority voting across candidate solutions from multiple LLMs, which mitigates model-specific biases and enhances generalization, thereby yielding a reliable test output for each input. Step 2 then identifies the top-performing candidate solution by incorporating test case difficulty weighting alongside a hold-out validation set.

\textit{Step 1: Verification of Test Cases via Consensus Voting.} 
First, we establish a preliminary ground truth for each test case input. For a given input $x_i$, we execute all candidate solutions to obtain a set of outputs $\{y^1_i, y^2_i, \ldots, y^m_i\}$, where $y^j_i = A^j(x_i)$. A provisional ground truth output $\hat{y}_i$ is determined via majority voting:
\begin{equation}
\hat{y}_i = \underset{y}{\operatorname{argmax}} \sum_{j=1}^{m} \mathbb{I}(y^j_i = y) \quad ,
\end{equation}
where $\mathbb{I}(\cdot)$ is the indicator function. Empirical evaluation on the TACO dataset demonstrates that this voting strategy achieves 94.7\% labeling accuracy with 8 sampled solutions (see analysis in Appendix~\ref{appendix_sec:validation-effectiveness} and Table~\ref{tab:labeling_accuracy}). This yields a candidate test set $\mathcal{T}_{candidate} = \{(x_1, \hat{y}_1), \ldots, (x_n, \hat{y}_n)\}$. Crucially, we posit that not all test cases are of equal importance; boundary or edge cases are critical for robust evaluation. We therefore introduce a weighting function $w(x_i) \to w_i$ that assigns a higher score to more challenging test cases. The weight $w_i$ is determined by combining semantic-based heuristics (e.g., assigning higher weights to boundary and stress tests explicitly requested in the prompt) and size-based metrics (e.g., input file size as a proxy for computational complexity). Detailed weighting criteria are provided in Appendix~\ref{appendix_sec:Test-Case Weighting Criteria}.

\textit{Step 2: Verification of Solutions via Weighted Evaluation and Hold-out Validation.} 
For reliable golden-solution selection, we use 50\% of $\mathcal{T}_{candidate}$ as the hold-out validation set $\mathcal{T}_{val}$ and the remainder as the weighted test suite $\mathcal{T}_{golden}$.

Each $A^j$ is scored on $\mathcal{T}_{golden}$.

% \begin{equation}
% A'_{golden} = \underset{A^j \in \{A^1, \ldots, A^m\}}{\operatorname{argmax}} \sum_{(x_i, \hat{y}_i) \in \mathcal{T}_{golden}} w_i \cdot \mathbb{I}(A^j(x_i) = \hat{y}_i) \quad .
% \end{equation}

\begin{equation}
\scalebox{0.80}{$ 
A'_{golden} = \underset{A^j \in \{A^1, \ldots, A^m\}}{\operatorname{argmax}} \sum_{(x_i, \hat{y}_i) \in \mathcal{T}_{golden}} w_i \cdot \mathbb{I}(A^j(x_i) = \hat{y}_i)
$}
\label{eq:2}
\end{equation}

The final confirmation of $A_{golden}$ depends on its performance on the unseen hold-out set $\mathcal{T}_{val}$.
We verify that $A'_{golden}$ also achieves the highest (or a competitively high) unweighted accuracy on $\mathcal{T}_{val}$ relative to other candidates. This additional validation step ensures that the selected solution is not merely overfitted to the specifics of the weighted test cases but demonstrates superior, generalizable correctness. The detailed algorithm is provided in Appendix~\ref{appendix_sec:dual-verification-algorithm}.

We ensure task solvability by using GPT-5 (high reasoning effort) as a proxy solver, filtering out approximately 36.9\% of tasks where it achieves a zero pass rate on the voted test cases. We posit that if such a capable model fails completely, the task likely suffers from ambiguity or underspecification. In a human audit of 150 randomly sampled SFT tasks, 90.0\% were well-defined, while golden-solution correctness and voted test-label accuracy were both 87.3\%; Appendix~\ref{appendix_sec:human-audit} reports the protocol, residual noise, and filtering fix. The pass-rate distribution of the remaining tasks is shown in Table~\ref{tab:prop_llm_pass_distribution} (Appendix~\ref{appendix_sec:task-solvability-analysis}).

Ultimately, this process yields a verified solution $A_{golden}$ and a comprehensive test suite $\mathcal{T}_{golden}$ for every generated task $q$. We leverage these high-quality synthetic assets to drive post-training for Code LLMs: $(q, A_{golden})$ pairs for supervised fine-tuning, and $(q, \mathcal{T}_{golden})$ for reinforcement learning via GRPO algorithm~\citep{shao2024deepseekmath}.

\section{Experiment}
\noindent
\textbf{Data and Training.}
We adopt GPT-o3-mini~\citep{openai2025o3mini} for task formulation, DeepSeek-R1-0528~\citep{deepseekai2025deepseekr1incentivizingreasoningcapability} and Qwen3-235B-A22B-Thinking-2507~\citep{yang2025qwen3technicalreport} for solution sampling, and R1-0528 for test case generation. 
Statistics for SFT datasets are provided in Appendix~\ref{appendix_sec: SFT Dataset Statistics}.
For SFT, we set the learning rate at 5e-5, with a global batch size of 128 for 8 epochs. 
For RL, reward is the fraction of passed tests (Appendix~\ref{appendix_sec:reward_function}).
The program executes in an isolated Redis sandbox for concurrent code testing (Appendix~\ref{appendix_sec: rl infra}).
Training configurations and costs are supplemented in Appendix~\ref{appendix_sec: time cost}.

\begin{table*}[!t]
\centering
\small
\setlength{\tabcolsep}{4pt}
\caption{Cross-paper overview on LiveCodeBench. Metrics, base models, inference settings, and data sizes follow the source papers or model cards and are not directly comparable; Table~\ref{tab:ocr_vs_synthsmith} provides a controlled comparison, and Appendix Table~\ref{tab:unified_public_eval} reevaluates public checkpoints under avg@8. $\dagger$: OpenThinker3 and rStar-Coder mix real and synthetic data. \method~uses only synthetic training artifacts.}
\label{tab:main_results}
\resizebox{\textwidth}{!}{%
\begin{tabular}{l|lcccccc|c|c}
\toprule
\textbf{Model} & \textbf{Base Model} & \textbf{SFT} & \textbf{RL} & \textbf{Size} & \textbf{Data} & \textbf{Task} & \textbf{Metric} & \textbf{LCB v5} & \textbf{LCB v6} \\
\midrule
\multicolumn{10}{c}{\textbf{SFT Approaches}} \\
Bespoke-Stratos~\citep{bespoke_stratos} & Qwen2.5-Instruct~\citep{qwen2025qwen25technicalreport} & \cmark & \xmark & 7B & 17k & Real & pass@1 & 16.2 & 8.57 \\
Qwen2.5-Coder-7B-Instruct~\citep{hui2024qwen25codertechnicalreport} & Qwen2.5-Coder-7B-Base & \cmark & \xmark & 7B & - & Real & - & 17.5 & 21.7 \\
Qwen2.5-Coder-14B-Instruct~\citep{hui2024qwen25codertechnicalreport} & Qwen2.5-Coder-14B-Base & \cmark & \xmark & 14B & - & Real & - & 24.0 & 23.4 \\
OpenThinker3~\citep{guha2025openthoughtsdatarecipesreasoning} & Qwen2.5-Instruct & \cmark & \xmark & 7B & 1.2m & Mixed$^\dagger$ & - & 51.7 & 40.8 \\
OlympicCoder~\citep{openr1} & Qwen2.5-Coder-Instruct~\citep{hui2024qwen25codertechnicalreport} & \cmark & \xmark & 7B & 100k & Real & - & 40.9 & 19.3 \\
OCR-Qwen-Instruct~\citep{ahmad2025opencodereasoningadvancingdatadistillation} & Qwen2.5-Instruct & \cmark & \xmark & 7B & 736k & Real & avg@64 & 51.3 & 44.5 \\
rStar-Coder~\citep{liu2025rstarcoderscalingcompetitivecode} & Qwen2.5-Coder-Instruct & \cmark & \xmark & 7B & 580K & Mixed$^\dagger$ & avg@16 & 57.3 & -- \\
Qwen3-8B~\citep{yang2025qwen3technicalreport} & Qwen3-8B-Base & \cmark & \xmark & 8B & - & Real & - & 57.5 & 48.4 \\
\multicolumn{10}{c}{\textbf{RL Approaches}} \\
Skywork-OR1~\citep{he2025skywork} & R1-Distilled-Qwen~\citep{deepseekai2025deepseekr1incentivizingreasoningcapability} & \xmark & \cmark & 7B & 124k & Real & avg@32 & 47.6 & 40.0 \\
DeepCoder-Preview~\citep{deepcoder2025} & R1-Distilled-Qwen & \xmark & \cmark & 14B & 24k & Real & pass@1 & 57.9 & 48.5 \\
AReal-boba²~\citep{fu2025areal} & R1-Distilled-Qwen & \xmark & \cmark & 14B & 24k & Real & avg@32 & 58.1 & 56.7 \\
\multicolumn{10}{c}{\textbf{SFT-then-RL Approaches (Stage 1)}} \\
AceReason1.1-SFT~\citep{liu2025acereasonnemotron11advancingmath} & Qwen2.5-Math~\citep{yang2024qwen25mathtechnicalreportmathematical} & \cmark & \xmark & 7B & 2.2M  & Real & avg@8 & 51.2 & - \\
MiMo-SFT~\citep{xiaomi2025mimounlockingreasoningpotential} & MiMo-Base & \cmark & \xmark & 7B & 500k  & Undisclosed & avg@8 & 52.3 & 45.5 \\
Klear-Reasoner-SFT~\citep{su2025klearreasoneradvancingreasoningcapability} & Qwen3-Base~\citep{yang2025qwen3technicalreport} & \cmark & \xmark & 8B & 1.5m  & Real & avg@8 & 58.5 & 49.6 \\
\blue{\method-Qwen2.5-7B-SFT} & \blue{Qwen2.5-Coder-7B-Instruct}  & \blue{\cmark} & \blue{\xmark} & \blue{7B} &  \blue{200k} & \blue{\textbf{Syn}} & \blue{avg@8} &  \blue{$\textbf{60.3}_{\pm 2.5}$} & \blue{$\textbf{53.5}_{\pm 1.7}$} \\
\blue{\method-Qwen2.5-14B-SFT} & \blue{Qwen2.5-Coder-14B-Instruct}  & \blue{\cmark} & \blue{\xmark} & \blue{14B} &  \blue{200k} & \blue{\textbf{Syn}} & \blue{avg@8} & \blue{$\textbf{64.2}_{\pm 1.7}$} & \blue{$\textbf{59.1}_{\pm 2.5}$} \\
\multicolumn{10}{c}{\textbf{SFT-then-RL Approaches (Stage 2)}} \\
AceReason1.1 & AceReason1.1-SFT & \cmark & \cmark & 7B & -  & Real & avg@8 & 57.2 & 52.1 \\
MiMo & MiMo-SFT & \cmark & \cmark & 7B & 130k  & Undisclosed & avg@8 & 57.8 & 49.3 \\
Klear-Reasoner & Klear-Reasoner-SFT & \cmark & \cmark & 8B & 106k  & Real & avg@8  & 61.6 & 53.1 \\
\blue{\method-Qwen2.5-7B-RL} & \blue{\method-Qwen2.5-7B-SFT} & \blue{\cmark} & \blue{\cmark} & \blue{7B} & \blue{40k}  & \blue{\textbf{Syn}} & \blue{avg@8} & \blue{$\textbf{62.9}_{\pm 1.8}$} & \blue{$\textbf{55.8}_{\pm 1.9}$} \\
\blue{\method-Qwen2.5-14B-RL} & \blue{\method-Qwen2.5-14B-SFT} & \blue{\cmark} & \blue{\cmark} & \blue{14B} & \blue{40k}  & \blue{\textbf{Syn}} & \blue{avg@8} & \blue{$\textbf{67.5}_{\pm 1.2}^{}$} & \blue{$\textbf{63.4}_{\pm 2.6}^{}$} \\
\bottomrule
\end{tabular}%
}
\end{table*}

\begin{figure*}[!t]
    \centering
    \includegraphics[width=1\linewidth]{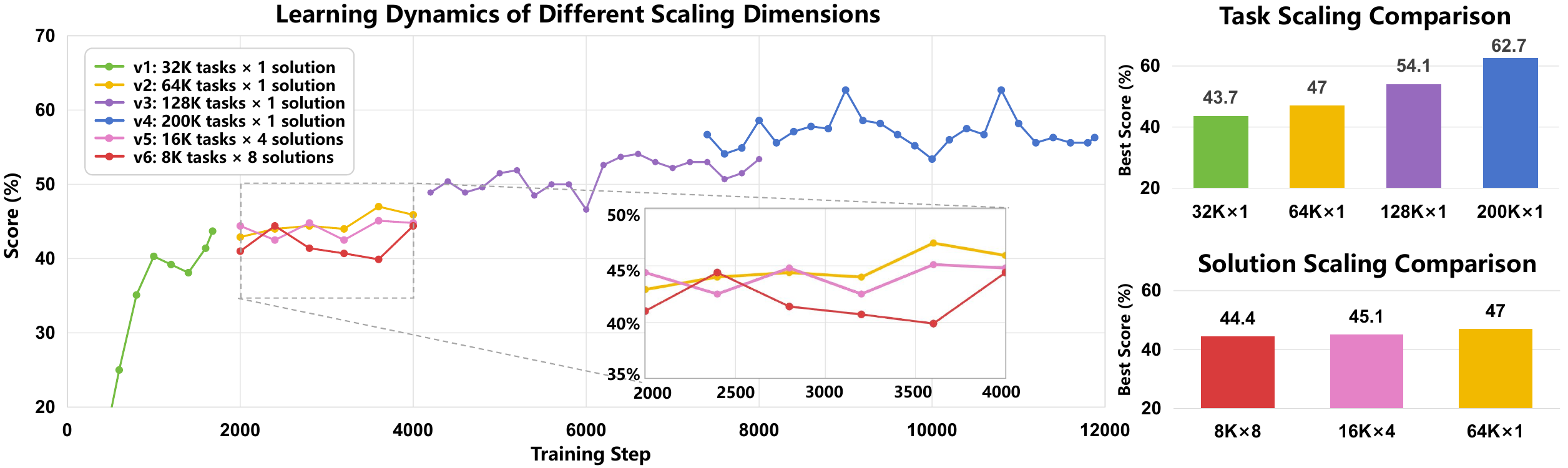}
    \caption{Left: Data scaling on LiveCodeBench v5. Right: Scaling unique tasks and scaling solutions per task.}
    \label{fig:scaling}
\end{figure*}

\noindent \textbf{Evaluation.}
We evaluate X-Coder on LiveCodeBench v5 and v6~\citep{jain2024livecodebenchholisticcontaminationfree}. Avg@8 is the mean pass rate over 8 independent completions per problem. We use temperature 0.6, top-p 0.95, top-k 20, and a maximum generation length of 32,768 tokens. We further examine whether synthetic training transfers across model families and benchmarks in Section~\ref{sec:generality}.

\subsection{Main Results}
As shown in Table~\ref{tab:main_results}, during the SFT stage, X-Coder achieves avg@8 scores of 60.3/53.5 on LiveCodeBench v5/v6 with the 7B model, and 64.2/59.1 with the 14B model. After RL, the scores improve to 62.9/55.8 and 67.5/63.4.

\subsection{SFT Experiments and Analysis}
\noindent \textbf{Scaling Laws Hold for Synthetic Data.} During the SFT stage, we investigate a central question: Can the SFT dataset be effectively scaled, and along which dimension should it be scaled more favorably?
To explore this, we are inspired by AceReason-Nemotron 1.1~\citep{liu2025acereasonnemotron11advancingmath} and expand the SFT dataset from two distinct perspectives: increasing the number of unique tasks and enlarging the number of solutions per task.
We design six subsets (v1--v6): v1--v4 increase the number of unique tasks (32k, 64k, 128k, and 200k unique prompts, each with 1 solution), while v5--v6 expand the number of solutions per task (16k unique prompts with 4 solutions, and 8k unique prompts with 8 solutions). The results in Figure~\ref{fig:scaling} reveal a promising scaling trend: performance improves monotonically as the dataset size increases (v1 $\to$ v4), rising steadily from 43.7\% to 62.7\%.

\noindent \textbf{Task Diversity Matters More than Solution Diversity.} The performance hierarchy v2 (64k$\times$1) $>$ v5 (16k$\times$4) $>$ v6 (8k$\times$8) shows that, under a fixed compute budget, scaling the number of unique tasks is consistently more effective than increasing solutions per task. This suggests that broader algorithmic coverage across diverse problems drives stronger generalization than repeated exposure to variations of the same problem.

\begin{table*}[!ht]
\centering
\small
\caption{Head-to-head Comparison with OpenCodeReasoning on LCB v5 using Qwen2.5-Coder-7B-Instruct.}
\label{tab:ocr_vs_synthsmith}
\begin{tabular}{lccccc}
\toprule
\textbf{Model} & \textbf{Avg. Pass@1} & \textbf{Pass@8} & \textbf{Easy} & \textbf{Medium} & \textbf{Hard} \\
\midrule
OpenCodeReasoning-Qwen (28k prompts x $\sim$26 solutions) & 53.6 & 69.0 & 95.2 & 67.0 & 21.8 \\
X-Coder-7B-SFT (25k prompts x 8 solutions) & 52.5 & 74.9 & 82.0 & 67.4 & 27.7 \\
X-Coder-7B-SFT (200k prompts x 1 solution) & \textbf{60.3} & \textbf{77.0} & \textbf{96.8} & \textbf{73.3} & \textbf{37.8} \\
\bottomrule
\end{tabular}
\end{table*}

\noindent \textbf{Synthetic Data Offers a Strong Alternative.}
We conduct a token-matched comparison with OpenCodeReasoning~\citep{ahmad2025opencodereasoningadvancingdatadistillation}, the largest reasoning dataset built upon real-world competitive programming tasks. As shown in Table~\ref{tab:ocr_vs_synthsmith}, our 25k$\times$8 setting is slightly weaker than OpenCodeReasoning in the overall score. This is mainly because it uses fewer unique prompts and allocates more tokens to multiple solutions of the same tasks, leading to weaker coverage of easy and common problem patterns.

However, this setting already performs better on the medium and hard splits, suggesting that our synthetic tasks provide effective supervision for more challenging reasoning. When we scale along the more effective dimension, namely increasing the number of unique prompts, X-Coder-7B-SFT with 200k$\times$1 data outperforms OpenCodeReasoning by 6.7 points. This result is consistent with our scaling analysis, which shows that task diversity matters more than the number of solutions per task. Synthetic artifacts reduce near-duplicate contamination risk (Appendix~\ref{appendix_sec:data_leakage}), but seed features and teachers retain real-data dependence.

\subsection{RL Experiments and Analysis}

After RL, X-Coder-14B improves from 64.2/59.1 to 67.5/63.4 on LiveCodeBench v5/v6, and the 7B model improves from 60.3/53.5 to 62.9/55.8.

Code reinforcement is a complementary stage after SFT. While SFT accounts for most of the performance jump, RL consistently extracts additional gains from execution rewards.

\begin{figure}[!t]
    \centering  
    \includegraphics[width=0.48\textwidth]{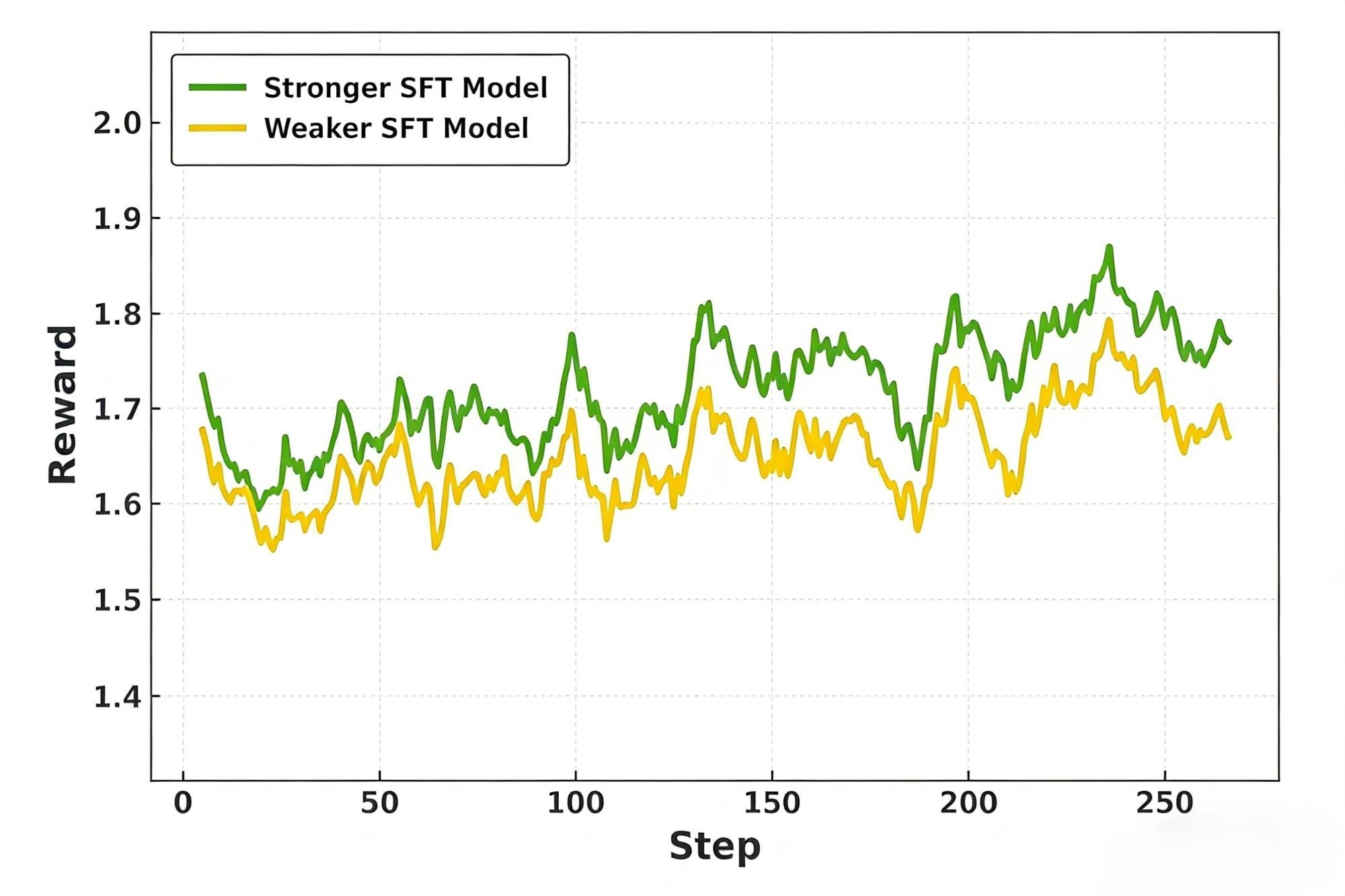}
    \caption{Reward dynamics of different SFT models.}
\label{fig:weak_and_strong}
\end{figure}

\noindent \textbf{SFT Absorbs Most Gains, but RL Still Provides Non-trivial Improvements.} Table~\ref{tab:model_generality} decomposes the contributions of SFT and RL across Qwen Instruct, Qwen Base, and Llama backbones. A consistent pattern emerges: SFT provides the dominant first-order improvement, while RL further improves the already strong SFT models. This suggests that verified long-CoT supervision captures most of the performance gain, but does not exhaust the optimization potential of the model. RL should not be viewed as a minor polishing step. Instead, it serves as a complementary optimization stage that uses executable feedback to refine behaviors that are difficult to fully learn from imitation alone.

\noindent \textbf{Strong SFT Initializers Enable More Effective RL.} The reward dynamics in Figure~\ref{fig:weak_and_strong} further support this view, showing that stronger SFT initializers lead to more effective RL optimization. Starting from a stronger SFT model places the policy in a more promising region of the solution space, making executable rewards more informative and easier to exploit. With the same RL setup, the stronger SFT initializer earns higher rewards.

\section{Generality}
\label{sec:generality}

We test whether gains from synthetic training artifacts transfer beyond one model and benchmark.

\noindent \textbf{Model Generality.} We apply the same pipeline to Qwen2.5-Coder-7B-Instruct, Qwen3-8B-Base, Llama-3.1-8B-Instruct, and \mbox{Qwen3.5-4B}\footnote{\nolinkurl{huggingface.co/Qwen/Qwen3.5-4B}.}. Synthetic training consistently transfers across model families and generations (Table~\ref{tab:model_generality}). \mbox{Qwen3.5-4B (32k)} improves from 39.7/37.0 to 65.6/58.8 after SFT and RL; at 80k, it gains 4.2/6.0 points over the original on v5/v6.

\begin{table}[!t]
\centering
\small
\setlength{\tabcolsep}{2pt}
\caption{Avg@8 results of SFT (200k) and RL (40k) across backbones. Parentheses denote the inference-token budget. The best result in each column is bold.}
\label{tab:model_generality}
\resizebox{\columnwidth}{!}{%
\begin{tabular}{@{}lcccccc@{}}
\toprule
\multirow{2}{*}{\textbf{Model}} & \multicolumn{3}{c}{\textbf{LCB v5}} & \multicolumn{3}{c}{\textbf{LCB v6}} \\
\cmidrule(lr){2-4} \cmidrule(lr){5-7}
 & \textbf{Ori.} & \textbf{SFT} & \textbf{RL} & \textbf{Ori.} & \textbf{SFT} & \textbf{RL} \\
\midrule
Qwen2.5-Coder-7B-Instruct & 17.5 & 60.3 & 62.9 & 21.7 & 53.5 & 55.8 \\
Qwen3-8B-Base & 16.4 & 59.4 & 64.0 & 15.4 & \textbf{55.4} & 56.5 \\
Llama-3.1-8B-Instruct & 11.8 & 42.7 & 47.8 & 14.3 & 35.9 & 39.4 \\
Qwen3.5-4B (32k) & 39.7 & 63.0 & 65.6 & 37.0 & 54.6 & 58.8 \\
Qwen3.5-4B (80k) & \textbf{63.1} & \textbf{63.2} & \textbf{67.3} & \textbf{55.8} & 55.3 & \textbf{61.8} \\
\bottomrule
\end{tabular}%
}
\end{table}

\noindent \textbf{Benchmark Generality.} To assess whether \method~generalizes beyond LiveCodeBench, we evaluate on LiveBench~\citep{white2025livebench}, Codeforces~\citep{penedo2025codeforces}, and MBPPPlus~\citep{evalplus}. As shown in Table~\ref{tab:benchmark_generality}, \method~achieves the highest average score across all three benchmarks, confirming broadly transferable code reasoning ability.

\begin{table}[!t]
\centering
\small
\setlength{\tabcolsep}{2pt}
\caption{Avg@8 results across LiveBench, Codeforces, and MBPP+.}
\label{tab:benchmark_generality}
\resizebox{\columnwidth}{!}{%
\begin{tabular}{@{}p{0.48\columnwidth}cccc@{}}
\toprule
\textbf{Model} & \textbf{LiveBench} & \textbf{Codeforces} & \textbf{MBPP+} & \textbf{Avg} \\
\midrule
X-Coder-14B-RL & \textbf{82.7} & \textbf{48.3} & 73.0 & \textbf{68.0} \\
X-Coder-14B-SFT & 75.5 & 45.6 & 73.4 & 64.8 \\
X-Coder-7B-RL & 70.3 & 38.1 & 71.8 & 60.1 \\
X-Coder-7B-SFT & 68.0 & 38.4 & 72.7 & 59.7 \\
AceReason1.1-7B & 68.2 & 32.8 & \textbf{75.0} & 58.7 \\
Klear-Reasoner-8B-SFT & 68.0 & 34.0 & 72.9 & 58.3 \\
MiMo-7B-RL & 66.7 & 35.1 & 70.7 & 57.5 \\
Qwen3-8B & 63.3 & 35.5 & 73.0 & 57.3 \\
MiMo-7B-SFT & 56.8 & 29.3 & 68.1 & 51.4 \\
OCR-Qwen-Instruct & 53.9 & 23.3 & 70.1 & 49.1 \\
Skywork-OR1-7B & 49.2 & 27.7 & 66.2 & 47.7 \\
DeepCoder-14B-Preview & 60.4 & 35.6 & 45.2 & 47.0 \\
OlympicCoder-7B & 38.3 & 18.4 & 64.9 & 40.5 \\
Qwen2.5-Coder-7B-Instruct & 39.6 & 6.6 & 62.7 & 36.3 \\
\bottomrule
\end{tabular}%
}
\end{table}

\section{Analysis}

\noindent \textbf{Dual-verification is Critical for Quality.} We employ a dual-verification strategy to mitigate noise in stochastic sampling. Table~\ref{tab:verification_comparison} confirms its efficacy: training on verified solutions (64k tasks, Qwen2.5-Coder-7B-Instruct) outperforms raw solutions by 6.4 points. Verifying 200k samples requires 1.6M CoT trajectories and 24M executions, but this is a one-time data-construction cost. Raw-solution training remains a lower-cost alternative~\citep{li2025llmseasilylearnreason,gandhi2025cognitive}. Detailed costs, human validation, and correlated-error analyses are provided in Appendix~\ref{appendix_sec:validation-effectiveness}.

\begin{table}[!b]
    \centering
    % \small
    \caption{Raw vs. verified solutions on LCB v5.}
    \label{tab:verification_comparison}
    \begin{tabular}{lcc}
        \toprule
        \textbf{Dataset} & \textbf{Size} & \textbf{LCB v5} \\
        \midrule
        Raw Solution & 64k & 47.0 \\
        Verified Solution & 64k & \textbf{53.4} \\
        \bottomrule
    \end{tabular}
\end{table}

\begin{figure}[!t]
\centering
\includegraphics[width=\columnwidth]{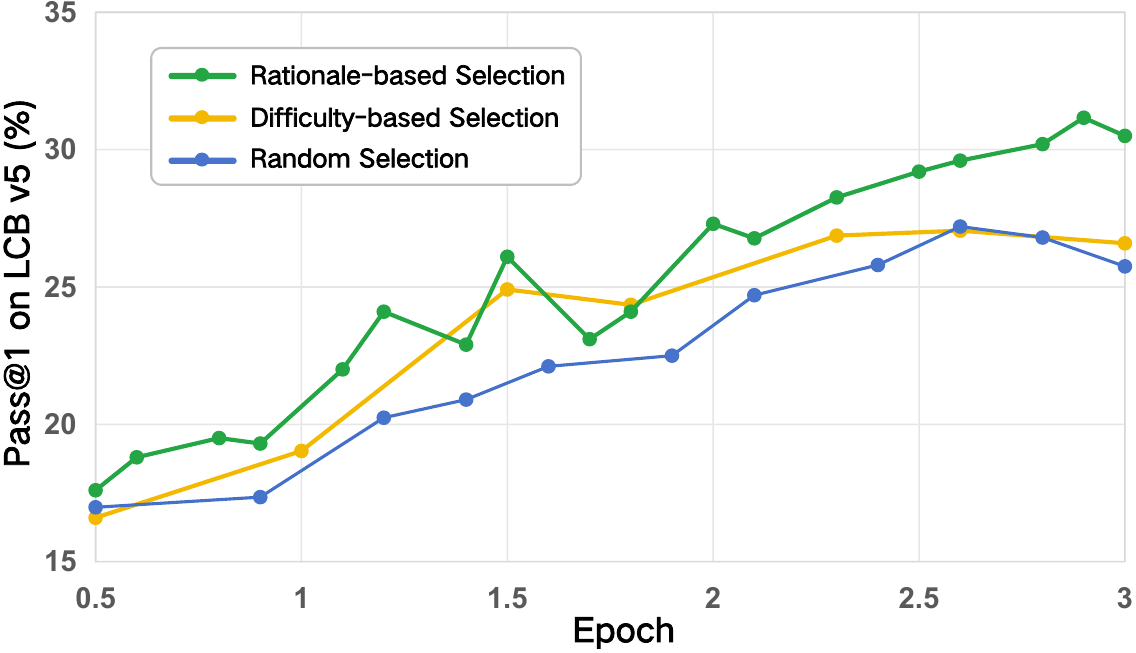}
\caption{Comparison of data selection.}
\label{fig:data_selection}
\end{figure}

\noindent \textbf{Long-CoT Tasks Provide a Practical Data Selection Signal.}
To investigate data utilization efficiency, we evaluate three 50k-task selection strategies from a 200k-task pool: difficulty-based selection using GPT-4o ratings, rationale-based selection using DeepSeek-R1-0528 reasoning length, and random selection. Solutions are generated by Qwen3-235B-A22B-Instruct-2507, and models are trained for 3 epochs. As shown in Figure~\ref{fig:data_selection}, tasks that induce longer CoT are more valuable training data for competitive programming, providing a practical heuristic for efficient data pruning in resource-constrained settings.

\noindent \textbf{The Long-CoT Tradeoff: Higher Quality, Slower Convergence.} CoT length matters: longer CoTs improve performance but raise training cost and slow convergence. We demonstrate this by training Qwen2.5-Coder-7B-Instruct on solutions generated by DeepSeek-R1-0528 (long-CoT) and Qwen3-235B-A22B-Instruct-2507 (short-CoT) for an identical set of 200k tasks.

\begin{table}[!t]
  \centering
  \small
  \caption{Long-CoT vs. short-CoT.}
  \label{tab:solution_type}
  \begin{tabular}{@{}lccc@{}}
    \toprule
    & \textbf{Epoch} & \textbf{LCB v5} & \textbf{LCB v6} \\
    \midrule
    \multirow{3}{*}{Short-CoT} & 3 & 35.0 & 29.3 \\
    & 8 & 43.1 & 37.6 \\
    & $\Delta$ & +8.1 & +8.3 \\
    \midrule
    \multirow{3}{*}{Long-CoT} & 3 & 42.9 & 36.0 \\
    & 8 & \textbf{60.3} & \textbf{53.5} \\
    & $\Delta$ & +17.4 & +17.5 \\
    \bottomrule
  \end{tabular}
\end{table}

As shown in Table~\ref{tab:solution_type}, the long-CoT approach achieves a 17.2\% absolute gain. This gain justifies the increased computational demand, which manifests as a slower convergence requiring 8--10 epochs compared to the 2--3 epochs needed for short-CoT data.

\begin{figure}[!t]
  \centering
  \includegraphics[width=\columnwidth]{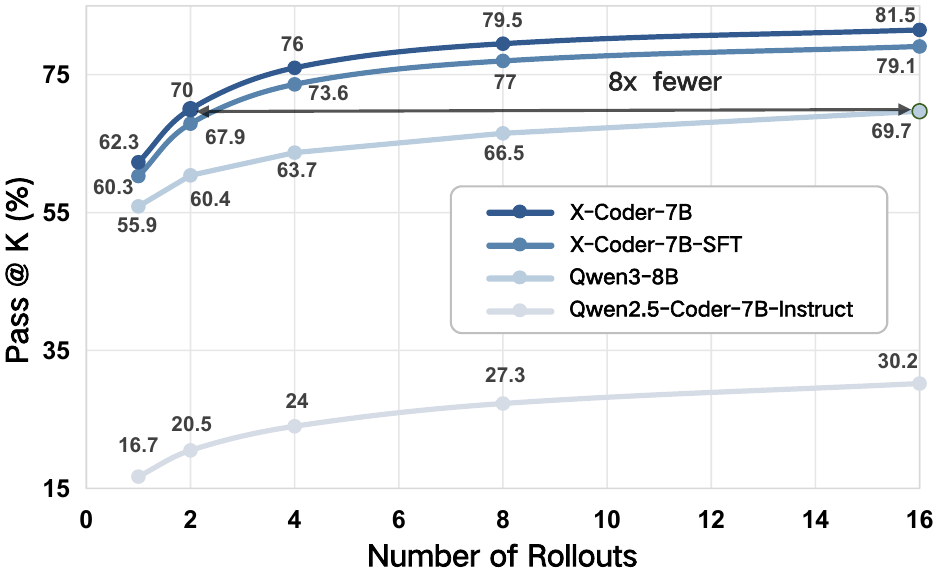}
  \caption{Test-time scaling performance.}
  \label{fig:tts}
\end{figure}

\noindent \textbf{Synthetic Data Improves Test Time Scaling.}
We compare the pass@k performance in Figure~\ref{fig:tts}. X-Coder-7B-RL improves over its foundation by 51.3 points in pass@16 and matches Qwen3-8B with 8$\times$ fewer rollouts. It also shows a larger pass@1-to-pass@16 gap than Qwen3-8B, suggesting more diverse reasoning trajectories.

\begin{table}[!t]
  \centering
  \caption{Distribution of failure cases on LCB v5. Base refers to Qwen2.5-Coder-7B-Instruct.}
  \label{tab:error_by_model}
  \small
  \begin{tabular}{@{}lccc@{}}
    \toprule
    \textbf{Error Type} & \textbf{Base} & \textbf{+SFT} & \textbf{+RL} \\
    \midrule
    Wrong Answer              & 194.6\textsubscript{±10.7} & 69.6\textsubscript{±3.7} & 67.9\textsubscript{±4.9} \\
    Context Exhaustion & 6.5\textsubscript{±8.2}  & 21.9\textsubscript{±3.7} & 11.8\textsubscript{±3.9} \\
    Time Limit Exceeded       & 18.1\textsubscript{±4.1}  & 13.7\textsubscript{±3.3} & 11.5\textsubscript{±2.6} \\
    Memory Limit Exceeded     & 0.0\textsubscript{±0.0}   & 0.0\textsubscript{±0.0}  & 0.17\textsubscript{±0.4} \\
    Signature Mismatch        & 0.0\textsubscript{±0.0}   & 0.0\textsubscript{±0.0}  & 1.0\textsubscript{±0.8} \\
    Syntax Error              & 0.0\textsubscript{±0.0}   & 0.0\textsubscript{±0.0}  & 8.3\textsubscript{±2.2} \\
    \bottomrule
  \end{tabular}
\end{table}

\noindent \textbf{Reasoning Errors Remain the Bottleneck.}
Table~\ref{tab:error_by_model} shows that wrong answers dominate across models, making algorithmic reasoning the main bottleneck. The next most frequent failures are context-window exhaustion and time limit errors. The former occurs when the model exhausts its 32k-token budget and emits no complete executable solution. Future work should target deeper reasoning, concise reasoning control, and execution efficiency.

Additional details on baselines, feature evolution, and qualitative reasoning behaviors are provided in Appendices~\ref{appendix_sec:baselines},~\ref{appendix_sec:Feature Extraction and Evolution}, and~\ref{appendix_sec: successful case}.

\vspace{2mm}
\section{Conclusion}
In this work, we demonstrate that synthetic training artifacts can support the complete SFT-to-RL cycle without directly using real-world tasks, solutions, or tests. Our pipeline produces solvable yet challenging tasks, verified long-CoT solutions, and executable tests, with dual verification reducing noise in selected solutions and test outputs. X-Coder-14B achieves 67.5\% avg@8 on LiveCodeBench v5 and 63.4\% on v6, outperforming its base model by over 40 points, while additional results transfer to newer and cross-family backbones. Our analysis highlights the value of task diversity, verified long-CoT supervision, and code-centric reinforcement learning. Our code and data are publicly available.

\section*{Limitations}

In this paper, we demonstrate that synthetic tasks, solutions, and tests can effectively support competitive-programming post-training. Our claim concerns these training artifacts, not the whole provenance chain: TACO-derived abstract features seed task synthesis, and teacher models may encode knowledge learned from real-world corpora. Frontier teachers are not strictly necessary (Appendix~\ref{appendix_sec:open-teacher}), but teacher quality affects downstream gains.

The limitations are twofold.
First, we focus on self-contained competitive-programming tasks. This setting captures core algorithmic reasoning and supports clean executable verification, but our conclusions may not directly extend to broader scenarios, such as repository-level generation and debugging~\citep{luo2026rpgrepositoryplanninggraph}, feature implementation, command-line environments, and visual coding tasks.

Second, dual verification is imperfect. Human auditing finds 12.7\% residual error in both golden solutions and voted labels before deterministic filtering, and correlated teacher errors can survive consensus. Lower-confidence samples are more tolerable in SFT than in RL, where wrong labels directly shape rewards; we therefore reserve more strongly verified data for RL. Test synthesis, feedback refinement, and adversarial testing may improve reliability and curb reward hacking.

\section*{Acknowledgements}
This work was partly supported by the National Natural Science Foundation of China (Grant No. 62576191), the Shenzhen Science and Technology Program (ZDCY20250901103533010).

% The generated bibliography is included as ordinary TeX so that arXiv does
% not need to run BibTeX or preserve a separately detected .bbl file.

\newpage
\appendix
\etocdepthtag.toc{mtappendix}
\etocsettagdepth{mtchapter}{none}
\etocsettagdepth{mtappendix}{subsection}

\setcounter{tocdepth}{3}
\renewcommand{\contentsname}{Appendix}
\renewcommand{\cftsecleader}{\cftdotfill{1}}
\renewcommand{\cftsecfont}{\normalfont}
\renewcommand{\cftsecpagefont}{\normalfont}
\renewcommand{\cftsecdotsep}{4.5}
\renewcommand{\cftsubsecdotsep}{4.5}
\renewcommand{\cftsubsecnumwidth}{2em}
\renewcommand{\cftsubsecindent}{1.5em}
\setlength{\cftbeforesecskip}{0.8em}
\setlength{\cftbeforesubsecskip}{0.25em}
\setlength{\cftbeforesubsubsecskip}{0.15em}

\ifcsname nolinenumbers\endcsname\nolinenumbers\fi
\etoctableofcontents
\thispagestyle{empty}

\newpage
\ifcsname nolinenumbers\endcsname\nolinenumbers\fi
\section{Training and Evaluation}
% \subsection{SFT-then-RL Training}
\label{appendix_sec: algorithm}

\noindent \textbf{Supervised Fine-tuning.} 
Given a dataset of task--solution pairs $\mathcal{D} = \{(x_i, y_i)\}_{i=1}^N$, the model with parameters $\theta$ is trained by minimizing the negative log-likelihood (NLL) of the target solution $y$ conditioned on the task $x$:
\begin{equation}
J_{\mathrm{SFT}}(\theta)
= - \mathbb{E}_{(x,y) \sim \mathcal{D}}
\Biggl[ \sum_{t=1}^{|y|} \log \pi_\theta \bigl(y_t \mid x, y_{<t}\bigr) \Biggr].
\end{equation}
The loss is applied over full long-CoT trajectories, including both reasoning steps and final code, enabling the model to imitate not only the solutions but also the underlying reasoning patterns.

\noindent \textbf{Reinforcement Learning.} Proximal Policy Optimization (PPO)~\citep{schulman2017proximalpolicyoptimizationalgorithms} is a widely adopted policy gradient method in Reinforcement Learning from Human Feedback (RLHF)~\citep{lambert2026reinforcementlearninghumanfeedback} for LLM due to its balance between exploration and exploitation and its empirical robustness. The method optimizes a policy $\pi_\theta$ by using a clipped surrogate objective to limit policy divergence, incorporating a value function to estimate expected rewards, and an entropy term to encourage exploration. The overall objective function for PPO is designed to maximize the policy performance while maintaining stability, and it is typically formulated as minimizing the following:
\newcommand{\pip}{\pi_\theta(a|s)}
\newcommand{\piold}{\pi_{\theta_{\mathrm{old}}}(a|s)}
\begin{equation}
\resizebox{0.98\columnwidth}{!}{$\displaystyle
\begin{aligned}
J_{\mathrm{PPO}}(\theta)
&= \mathbb{E}_{s \sim P(S), a \sim \pip} \biggl[ \min \biggl(
\frac{\pip}{\piold} A(s,a), \\
&\qquad \text{clip}\Bigl( \frac{\pip}{\piold}, 1-\epsilon, 1+\epsilon \Bigr) A(s,a)
\biggr) \biggr]
\end{aligned}
$}
\end{equation}
where the expectation is computed over states $s$ (drawn from distribution $P(S)$) and actions \( a \) (sampled from the current policy $\pi_\theta(a \mid s)$ ), combining the minimum of two terms: (1) the product of the probability ratio $\frac{\pi_\theta(a \mid s)}{\pi_{\theta_{\mathrm{old}}}(a \mid s)}$ and the advantage function $A(s, a)$, where the advantage function quantifies the relative benefit of taking action $ a$ in state $ s$; and (2) the same product but with the probability ratio clipped to the interval $[1 - \epsilon, 1 + \epsilon]$. Here, $\epsilon$ is a hyperparameter governing the magnitude of policy updates. This clipping mechanism effectively constrains excessive policy changes, thereby enhancing training stability.

\begin{figure*}[t]
  \centering
  \begin{subfigure}[t]{0.24\textwidth}
    \centering
    \includegraphics[width=\linewidth]{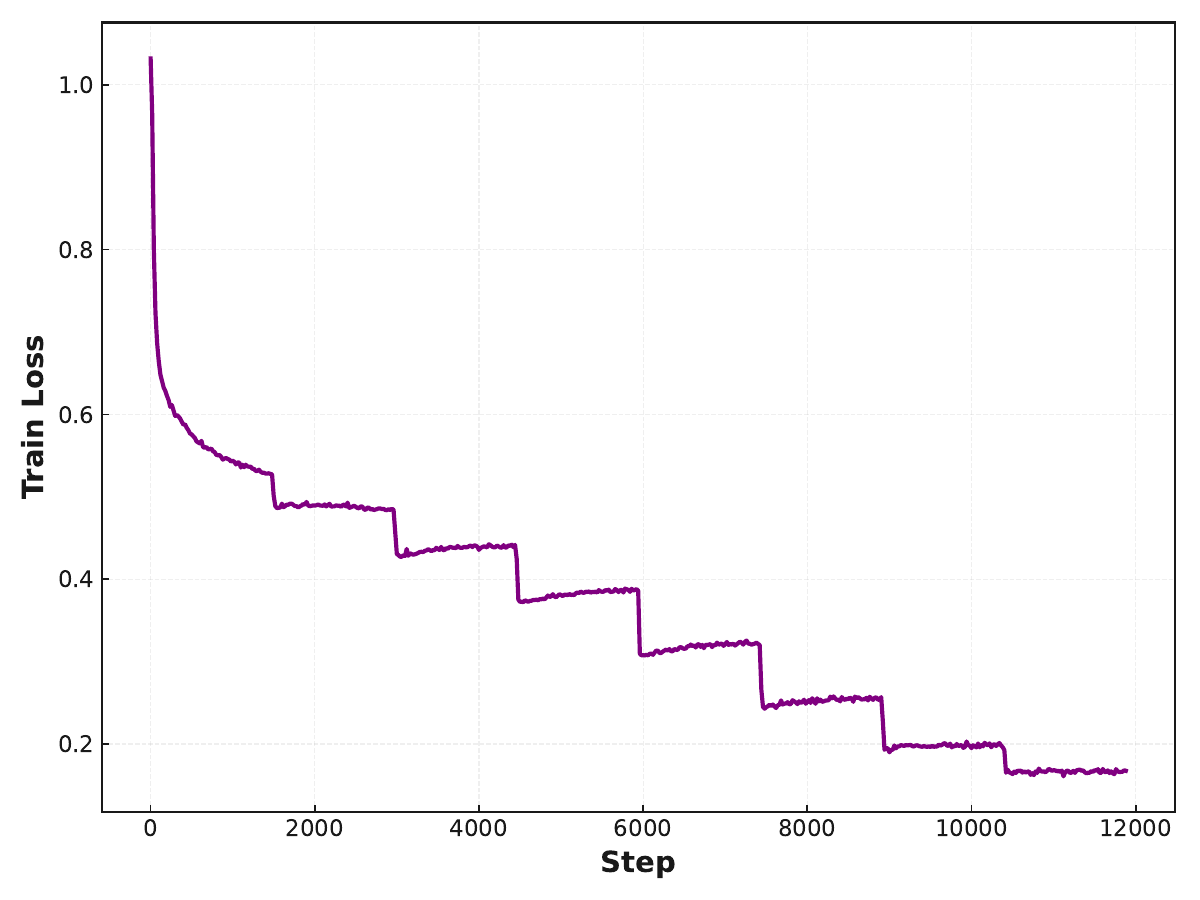}
    \caption{SFT loss.}
    \label{fig:train_loss_curve}
  \end{subfigure}\hfill
  \begin{subfigure}[t]{0.24\textwidth}
    \centering
    \includegraphics[width=\linewidth]{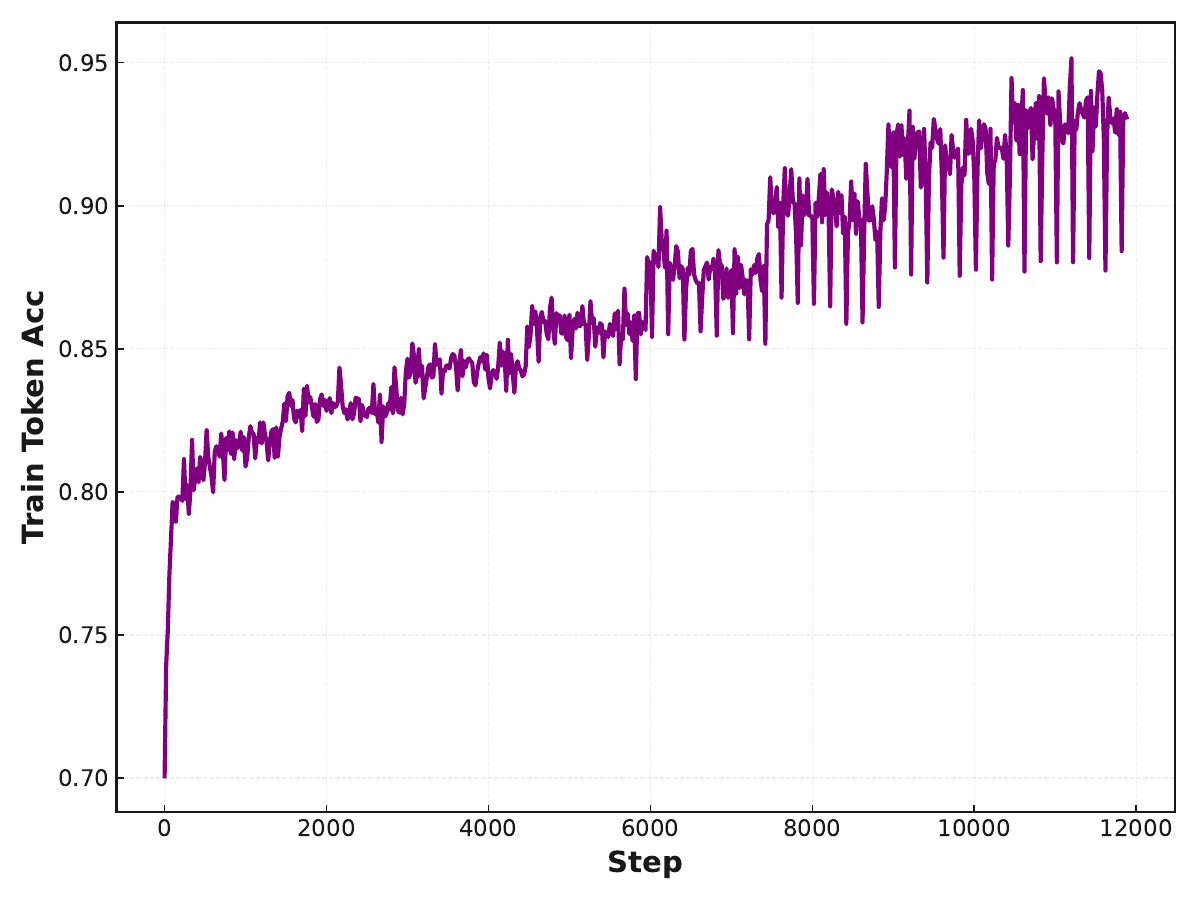}
    \caption{SFT token accuracy.}
    \label{fig:train_token_acc_curve}
  \end{subfigure}\hfill
  \begin{subfigure}[t]{0.24\textwidth}
    \centering
    \includegraphics[width=\linewidth]{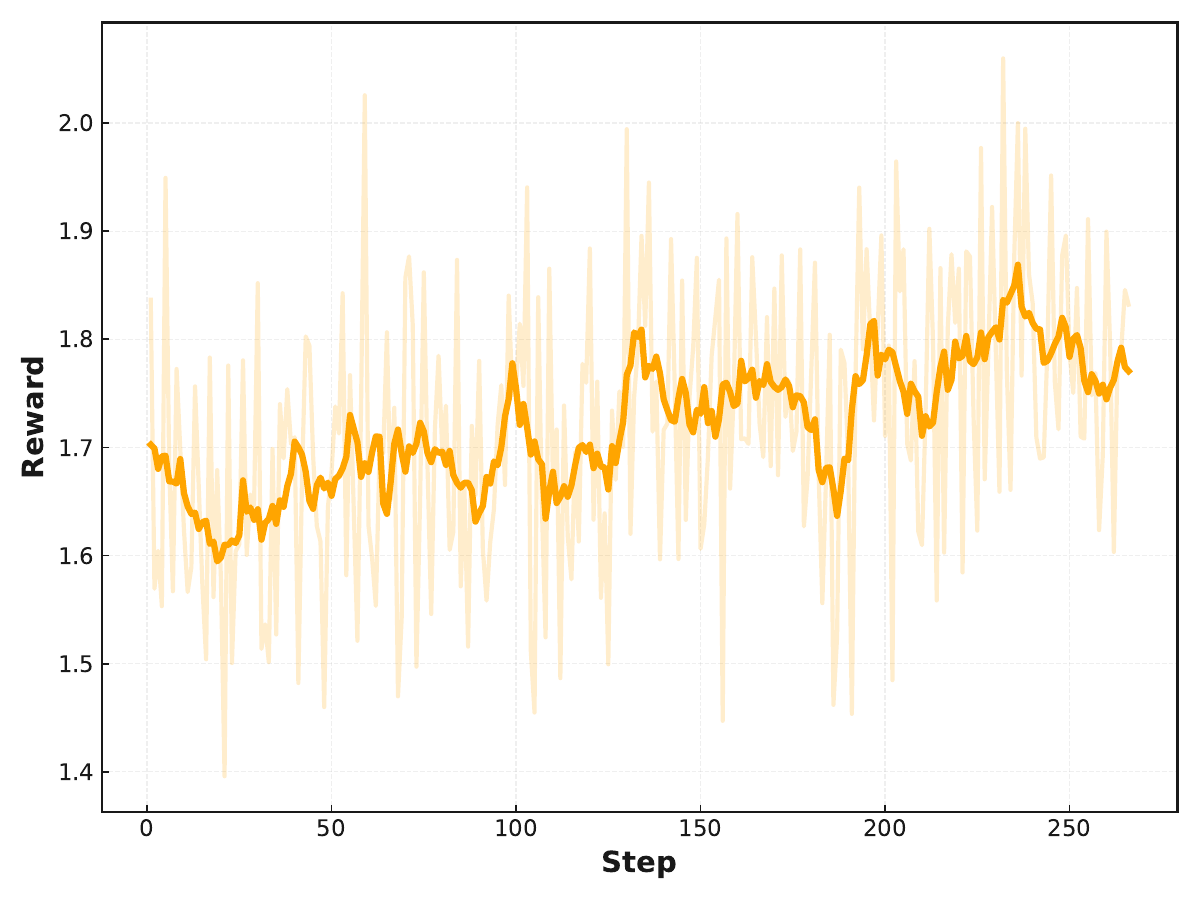}
    \caption{RL reward.}
    \label{fig:reward_curve}
  \end{subfigure}\hfill
  \begin{subfigure}[t]{0.24\textwidth}
    \centering
    \includegraphics[width=\linewidth]{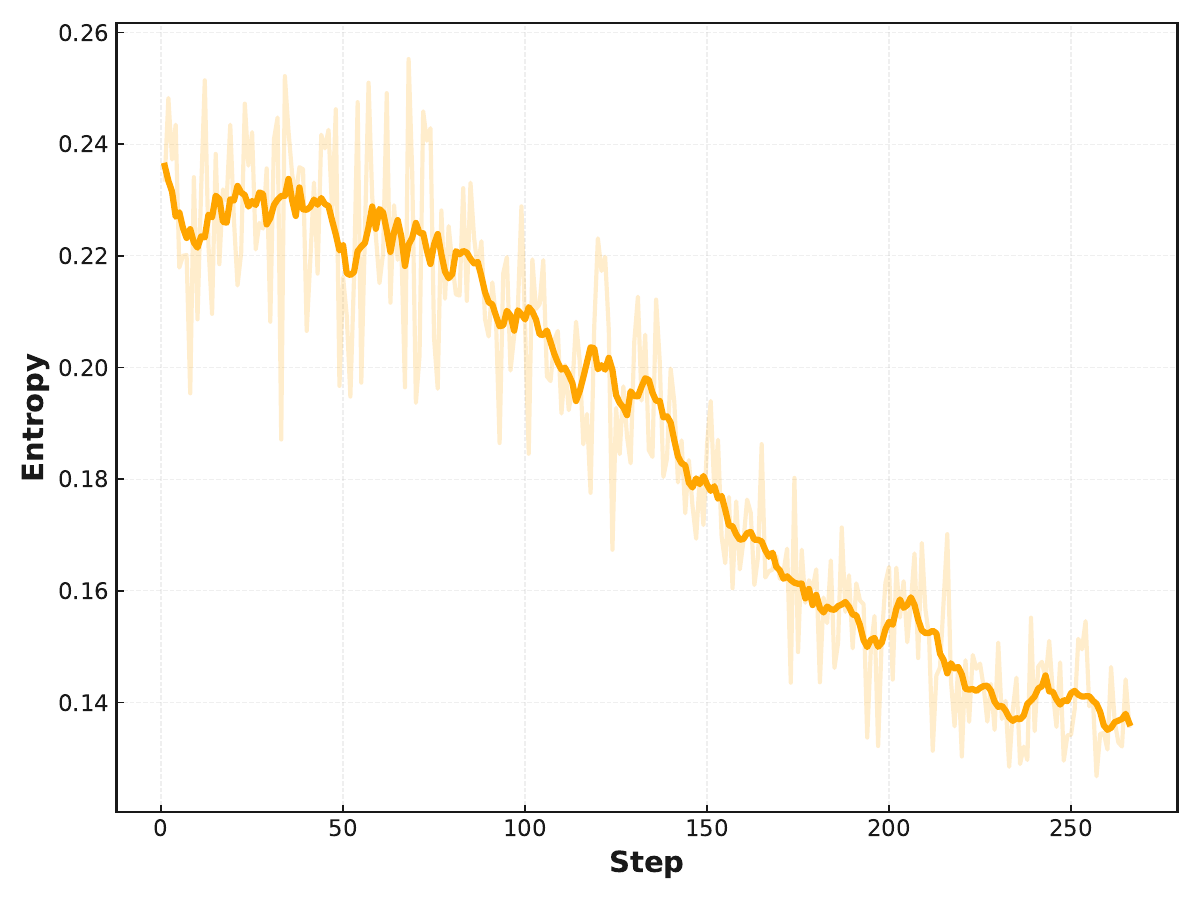}
    \caption{RL entropy.}
    \label{fig:entropy_curve}
  \end{subfigure}
  \caption{Training dynamics for SFT and RL.}
  \label{fig:training_dynamics}
\end{figure*}

\begin{figure*}[t]
    \centering
    \tikzstyle{block} = [rectangle, rounded corners, draw, text centered, minimum height=1.2cm, text width=3.1cm, font=\small]
    \tikzstyle{line} = [draw, -{Latex[length=2mm]}]
    \tikzstyle{biline} = [draw, {Latex[length=2mm]}-{Latex[length=2mm]}]

    \begin{tikzpicture}[node distance=1.4cm]
        \node[block] (api) {FastAPI Web \\ (API Gateway)};
        \node[block, right=of api] (redis) {Redis Queue \\ (Message Broker)};
        \node[block, right=of redis] (worker) {Worker Pool \\ (Code Executors)};

        \node[block, rounded corners=0, below=of api] (api-desc) {Request Handling};
        \node[block, rounded corners=0, below=of redis] (redis-desc) {Task \& Result Queues};
        \node[block, rounded corners=0, below=of worker] (worker-desc) {Sandbox Execution};

        % Draw arrows
        \path[biline] (api.east) -- (redis.west);
        \path[biline] (redis.east) -- (worker.west);
        \path[line] (api.south) -- (api-desc.north);
        \path[line] (redis.south) -- (redis-desc.north);
        \path[line] (worker.south) -- (worker-desc.north);
    \end{tikzpicture}
    \caption{The distributed architecture of the code verification framework.}
    \label{fig:arch-code-verification}
\end{figure*}

However, its application to LLMs encounters significant challenges, including substantial computational overhead from maintaining a critic network, which increases memory usage and training time for models with billions of parameters. To address these limitations, Group Relative Policy Optimization (GRPO)~\citep{shao2024deepseekmathpushinglimitsmathematical} has emerged as an efficient alternative. By eliminating the critic network, GRPO reduces computational and memory demands, estimating advantages directly from rewards of multiple rollouts to the same prompt, thus leveraging the comparative nature of reward models and offering a scalable solution for LLM training.
The GRPO objective function is mathematically formulated as an averaged composite expression across multiple rollouts, incorporating policy ratio optimization and KL regularization:
\newcommand{\rhoit}{\rho_{i,t}}
\newcommand{\defrhoit}{\rho_{i,t} = \dfrac{\pi_\theta(a_{i,t} | s, a_{i,<t})}{\pi_{\theta_{\mathrm{old}}}(a_{i,t} | s, a_{i,<t})}}
\begin{equation}
\resizebox{0.98\columnwidth}{!}{$\displaystyle
\begin{aligned}
J_{\mathrm{GRPO}}(\theta)
&= \frac{1}{G} \sum_{i=1}^G \frac{1}{|a_i|}
\sum_{t=1}^{|a_i|} \biggl\{ \min \biggl( \rhoit \hat{A}_{i,t}, \\
&\qquad \text{clip}\bigl( \rhoit, 1-\epsilon, 1+\epsilon \bigr) \hat{A}_{i,t} \biggr) \\
&\qquad - \beta D_{\mathrm{KL}}[\pi_\theta \parallel \pi_{\mathrm{ref}}] \biggr\}
\end{aligned}
$}
\end{equation}
where $\defrhoit$ denotes the probability ratio of the old and new strategies. $ G $ is the number of rollouts per prompt, $ |a_i| $ denotes the length of the $ i $-th action sequence, $ \hat{A}_{i,t} $ estimates the advantage of action $ a_{i,t} $ at timestep $ t $.  The clipping is analogous to PPO, and $ \beta $ penalizes deviations from $ \pi_{\text{ref}} $ via the KL-divergence term. The objective averages across rollouts and timesteps, combining a clipped probability ratio (to stabilize updates while leveraging advantage signals) with a KL penalty to balance policy improvement against alignment with the reference policy. This dual mechanism ensures controlled optimization by restricting drastic policy shifts while maintaining coherence with prior behavior.

\subsection{Reward Function.}
\label{appendix_sec:reward_function}
We remove formatting rewards (e.g., enforcing ``think" tags), as the SFT model already follows the format, allowing the policy to focus on passing test cases.
Given a rollout, the reward $R$ is practiced as:
\begin{equation}
\resizebox{\columnwidth}{!}{$\displaystyle
\mathcal{R} =
\begin{cases}
-2, & \text{if no code is extracted or the code fails to compile}, \\[6pt]
0, & \text{if the code compiles but passes no test cases}, \\[6pt]
\dfrac{5.0 \times \#\text{passed}}{\#\text{total}}, & \text{otherwise}.
\end{cases}
$}
\end{equation}
We adopt a continuous reward setting, as it provides denser supervision than the all-or-nothing alternative and leads to faster convergence~\citep{wei2025swerladvancingllmreasoning,DBLP:journals/corr/abs-2410-17621}.

\subsection{Training Dynamics.}
\label{appendix_sec: training dynamics}

As shown in Figure~\ref{fig:train_loss_curve} and Figure~\ref{fig:train_token_acc_curve}, we present the SFT training curves (loss and token accuracy). 
Figure~\ref{fig:reward_curve} and Figure~\ref{fig:entropy_curve} illustrate the RL training curves (reward and entropy).

\subsection{Training Configs and Costs}
\label{appendix_sec: time cost}
For SFT, we use a learning rate of 5e-5 with a global batch size of 128 for 8 training epochs. For RL, the policy models are updated with a global batch size of 512 and a consistent learning rate of 7e-5, without applying the KL-divergence constraint, and employ a rollout temperature of 1.0 with 8 rollouts for each query. Concretely, training X-Coder on Qwen2.5-Coder-7B-Instruct required 128 H20 Enterprise (96 GB) GPUs for 220 hours during SFT, and 32 H200 GPUs for 7 days to complete 270 update steps during RL.

Data construction is also compute-intensive: verifying 200k samples generates 1.6M long-CoT candidate trajectories and performs approximately 24M sandbox executions. This is a one-time construction expense; it improves LCB v5 by 6.4 points over training on raw solutions (Table~\ref{tab:verification_comparison}), while the raw variant remains a lower-cost option. We report trajectory and execution counts rather than a monetary API estimate because price schedules and deployment choices differ across teacher providers.

\subsection{Distributed Code Verification}
\label{appendix_sec: rl infra}
To provide a robust and scalable solution for code validation, we develop a distributed arbitration framework inspired by open-source repository implementations\footnote{\url{https://github.com/0xWJ/code-judge.git}}.  The system is based on a microservice architecture, comprising a \textit{FastAPI}-based asynchronous API Gateway, a pool of code execution workers in the sandbox and a central \textit{Redis} instance. \textit{Redis} serves as a high-performance message broker and state manager, effectively decoupling the client-facing gateway from the backend computational workers. This architectural choice facilitates independent scaling, deployment, and enhances the overall resilience of the system. 
\textbf{Based on this evaluation framework, we implemented highly concurrent code testing during RL training.} We used batching when submitting tasks to the \textit{Redis} server to achieve high concurrency even with low request rates. This process required the server to distribute all test tasks to different workers, utilizing the CPU power of all participating machines.
Figure~\ref{fig:arch-code-verification} shows the system diagram of the framework.

The framework's efficacy is derived from its strategic implementation of \textit{Redis} data structures. Task distribution is managed by a \textit{Sorted Set}, which functions as a time-prioritized FIFO queue; submissions are added with a timestamp score via \textit{ZADD}, and workers atomically retrieve the next task using \textit{BZPOPMIN}. This approach ensures ordered processing and prevents race conditions. For result transmission, each task is assigned a dedicated \textit{List}, to which a worker pushes the outcome using \textit{RPUSH}. The API Gateway then performs a blocking pop (\textit{BLPOP}) on this unique list to retrieve the corresponding result efficiently. Furthermore, worker health and presence are monitored using \textit{String} keys with a Time-To-Live (TTL). Workers periodically refresh their key's TTL as a heartbeat, enabling the system to automatically detect and de-register unresponsive nodes.

The resulting system exhibits several key advantages. The asynchronous, in-memory nature of its core components yields high throughput and low-latency performance. Its design is inherently scalable, as the stateless worker pool can be expanded horizontally to meet computational demand, while native support for \textit{Redis} Cluster addresses data-tier bottlenecks. Finally, the framework's reliability is bolstered by the atomicity of \textit{Redis} operations and the integrated fault-detection mechanism, ensuring dependable and consistent code verification.

\subsection{Baselines}
\label{appendix_sec:baselines}
We compare the \method~with three categories of baselines:  
(1) SFT model, e.g., Bespoke-Stratos, OlympicCoder, OCR-Qwen-Instruct, OpenThinker3, Qwen3-8B, and rStar-Coder;
(2) RL model, including Skywork-OR1, DeepCoder-14B-Preview, and AReal-boba²-14B;
(3) SFT-then-RL model, such as AceReason1.1, Klear-Reasoner, and MiMo-RL.

\subsection{Open-Source Teacher Ablation}
\label{appendix_sec:open-teacher}
To test whether the pipeline requires frontier proprietary teachers, we replace task formulation, test generation, and solution generation with Qwen3.5-35B-A3B, using neither GPT-5 filtering nor Qwen3-235B. We compare 32k tasks from this open-source variant against a 32k subsample of the proprietary pipeline, keeping the Qwen2.5-Coder-7B-Instruct student and SFT configuration fixed.

\begin{table}[!h]
\centering
\small
\caption{Teacher ablation on LCB v5 (avg@8).}
\label{tab:open_teacher}
\begin{tabular}{@{}lcc@{}}
\toprule
\textbf{Pipeline} & \textbf{Tasks} & \textbf{LCB v5} \\
\midrule
Base model & -- & 17.5 \\
Open-source teacher & 32k & 27.7 (+10.2) \\
Proprietary teachers & 32k & \textbf{35.9} (+18.4) \\
\bottomrule
\end{tabular}
\end{table}

Frontier teachers are therefore not necessary for the pipeline to produce substantial gains, although teacher quality determines the attainable improvement.

\begin{table*}[!t]
\centering
\small
\setlength{\tabcolsep}{8pt}
\caption{Public checkpoints reevaluated under a unified avg@8 protocol on LCB (8 samples, temperature 0.6, top-p 0.95, and official chat templates). Reported values retain each source's original metric and are shown only for context.}
\label{tab:unified_public_eval}
\begin{tabular}{@{}lcccc@{}}
\toprule
\textbf{Model} & \textbf{Reported v5} & \textbf{Reported v6} & \textbf{Avg@8 v5} & \textbf{Avg@8 v6} \\
\midrule
OlympicCoder-7B & 40.9 & -- & 41.2 & 34.1 \\
OCR-Nemotron-7B & 51.3 (avg@64) & 44.5 & 49.6 & 44.2 \\
Skywork-OR1-7B & 47.6 (avg@32) & 40.0 & 46.5 & 40.5 \\
DeepCoder-14B-Preview & 57.9 (pass@1) & 48.5 & 56.5 & 47.2 \\
\bottomrule
\end{tabular}
\end{table*}

% \newpage
\section{Novel Task Synthesis}
Building on EpiCoder, which synthesizes programming tasks through feature-based combinations, we introduce three key improvements to generate more diverse and complex instructions. 

\textit{First}, rather than relying on broad feature definitions, we explicitly extract and evolve competition-related features from 10,000 question--solution pairs in TACO~\citep{li2023taco} using GPT-4o-0513 (Appendix~\ref{appendix_sec:Feature Extraction and Evolution}). \textit{Second}, we adopt a two-stage process: selecting mutually consistent features and then formulating challenging, hint-free tasks (Appendix~\ref{appendix_sec:Stylized Task Generation for Competitive Programming}). \textit{Third}, we extend the synthesis method to support multi-style generation, covering CodeForces-style tasks (rich narratives with standard I/O), LeetCode-style tasks (starter code with fixed signatures), and AtCoder-style tasks (concise specifications), thereby enhancing task diversity. In Appendix~\ref{appendix_sec: Task Difficulty Estimates}, we further estimate the difficulty of synthesized problems using a trained discriminator.
 
\subsection{Feature Extraction and Evolution}
\label{appendix_sec:Feature Extraction and Evolution}
While EpiCoder extracts general-purpose features from raw corpus, we explicitly extract and evolve competitive programming-related feature. Specifically, we design multiple aspect of features that highly relates to competitive programming, such as data structure, algorithm, mathematical, etc.

We improve the extraction process to guide the LLM to focus on competitive programming--related concepts, as follows:
\begin{appendixjsonlisting}
Extract features from the provided problem and solution code related to algorithmic programming, competitive programming, Leetcode, and Codeforces, following the requirements for each category below, formatted in JSON structure.
Responses in the following categories should be concise and organized in a JSON format surrounded with <begin> and <end>. Categories may include nested structures if applicable. Here is an example of the expected format:
<begin>{
    "programming language": [
        "Python"
    ],
    "problem type": [
        "graph traversal"
    ],
    "algorithm": {
        "graph algorithms":[
            "Dijkstra's algorithm",
            "DFS",
            "BFS"
        ],
        "dynamic programming":[
            "Longest Increasing Subsequence",
            "Knapsack Problem"
        ]
    },
    ...
    "implementation logic":["recursive", "iterative"]
}<end>
Categories to extract:
1. Programming Language: Note the specific programming language used. Example: ["Python", "C++"].
2. Problem Type: Outline the type of problem the code is solving. Example: ["graph traversal", "sorting", "dynamic programming"].
3. Algorithm: Identify the specific algorithm or method being used in the code. This category can include the following subcategories:
    3.1 Graph Algorithms: Specify graph algorithms used. Example: ["Dijkstra's algorithm", "DFS", "BFS"].
    3.2 Sorting Algorithms: Specify sorting algorithms used. Example: ["QuickSort", "MergeSort"].
    ...
4. Data Structures: Describe the primary data structures utilized. Example: ["array", "graph", "tree", "heap"].
5. Implementation Logic: Describe the implementation logic. Example: ["iterative", "recursive", "bit manipulation"].
6. Complexity Analysis: Provide time and space complexity of the code if available. Example: ["Time Complexity: O(n log n)", "Space Complexity: O(n)"]
7. Optimization Techniques: Specify any optimizations applied. Example: ["memoization", "greedy approaches", "bitwise operations"].
...
Extract as many features as possible and try not to let a feature appear in multiple categories at the same time.
\end{appendixjsonlisting}

Then we increase the diversity and complexity through evolution along both the breadth and depth dimensions. For example, along the breadth dimension, given an extracted feature such as quicksort, the LLM may evolve new features like bubble sort, even if they were not originally extracted. Along the depth dimension, a concept such as prefix sum can evolve into more advanced variants like difference array or Fenwick tree, reflecting increasing levels of abstraction and difficulty.
The overall evolution process is illustrated below.

\newpage
\begin{appendixjsonlisting}
Feature Tree Evolution Task:
You are provided with a feature tree represented as a nested JSON structure. Each node in this tree represents a feature or a sub-feature of competitive algorithm programming, with the leaves being the most specific features. Your task is to expand this feature tree both in depth and breadth. Depth expansion means adding more specific sub-features to existing leaves. Breadth expansion means adding more sibling features at the current levels.

Here are some explanations of the features:
{explanations}
The input feature tree will be provided in JSON format, and your output should be a JSON structure that represents the expanded feature tree.
Output Format:
- Expanded Feature Tree: Provide the expanded feature tree as a JSON structure. Surround the json with <begin> and <end>.

Input Feature Tree Example:
{
    "algorithm": {
        "sorting": ["quick sort", "merge sort"],
        "tree traversal": ["in-order traversal"]
    },
    ...
}
Expanded Feature Tree Example:
<begin>
{
    "algorithm": {
        "sorting": {
            "quick sort": ["3-way quick sort", "dual-pivot quick sort"], 
            "merge sort": ["top-down merge sort", "bottom-up merge sort"], 
            "heap sort":[]
        },
        "tree traversal": {
            "in-order traversal": ["recursive in-order traversal", "iterative in-order traversal"],
            "pre-order traversal":[],
            "post-order traversal":[],
            "level-order traversal":[],
        }
    },
    ...
}
<end>
Constraints:
1. For breadth expansion, add at least 2 new sibling features to each existing node.
2. For deep expansion, you need to add new sub-features to it, provided that you think the current leaf node has a more fine-grained feature.
3. Focus on generating new and innovative features that are not present in the provided examples.
4. The features are related to competitive algorithm programming.
Please follow the above constraints and expand the feature tree accordingly.

Input:
{features}
Output:
<begin>expanded feature tree<end>
\end{appendixjsonlisting}

After evolution, we merge features that share common traits into a larger tree, providing a rich pool of features for subsequent task formulation.

\begin{table*}[!h]
    \centering
    \small
    \caption{Feature counts from a single-source TACO seed, a larger multi-source OpenCodeReasoning (OCR) corpus, and feature evolution. Evolution from 10k TACO tasks produces more features overall than extraction from 28k multi-source real tasks.}
    \label{tab:feature_evolution}
    \begin{tabular}{@{}lrrrr@{}}
        \toprule
        \textbf{Category} & \textbf{TACO 10k} & \textbf{OCR 28k} & \textbf{TACO Evolved} & \textbf{Growth} \\
        \midrule
        Algorithm & 27,400 & 168,274 & 176,914 & $\times 6.46$ \\
        Data Structures & 12,353 & 54,061 & 65,104 & $\times 5.27$ \\
        Problem Type & 14,134 & 73,481 & 130,293 & $\times 9.22$ \\
        Implementation Logic & 12,419 & 61,263 & 106,157 & $\times 8.55$ \\
        Complexity Analysis & 16,124 & 47,088 & 90,016 & $\times 5.58$ \\
        Optimization Techniques & 1,537 & 24,461 & 14,124 & $\times 9.19$ \\
        \midrule
        \textbf{Total} & \textbf{83,967} & \textbf{428,628} & \textbf{582,608} & $\times 6.94$ \\
        \bottomrule
    \end{tabular}
\end{table*}

\newpage
\subsubsection{Statistics for Feature Extraction and Evolution}
\label{appendix_sec:Statistics for Feature Extraction and Evolution}

We present detailed statistics on feature evolution and data filtering to demonstrate how the process expands feature diversity and yields a high-quality 240k dataset. The statistics of feature extracted and evolved are shown in Table~\ref{tab:feature_evolution}.

The evolution strategy is the primary driver of feature-pool breadth: starting from 10k TACO tasks, it expands individual categories by factors of 5.3--9.2 and yields 582,608 features, exceeding the 428,628 features directly extracted from the multi-source OCR corpus. Broader real-data seeds and evolution remain complementary: OCR provides a much stronger initial pool than raw TACO, and future versions can evolve multi-source seeds.

\subsection{Stylized Task Generation for Competitive Programming}
\label{appendix_sec:Stylized Task Generation for Competitive Programming}

We design a prompt template to systematically transform extracted features into stylized competitive programming tasks.  

\textbf{Input:} a sampled feature tree represented in JSON format. 

\textbf{Output:} a feature-role tree (JSON), where each node is assigned roles such as core technique, subroutine, or constraint, together with an integration strategy (string) that explains how to combine these features into a coherent problem.  

To improve instruction-following and task understanding, the template is equipped with a one-shot example that demonstrates how raw features are mapped into roles and integrated into a task.  

\newpage
\begin{appendixjsonlisting}
You are a professional competitive programming problem setter. Your task consists of three parts:

Step 1: Tree-Structured Feature Role Explanation
Recursively traverse the provided feature tree.
- For each leaf node, annotate it with a "potential_use" field describing how this feature is typically used in competitive programming problems (e.g., input modeling, optimization, search, handling edge cases, etc.).
- Internal nodes retain their structure for hierarchy.
Output the annotated tree in the same structure, with every leaf node containing its "potential_use".
Step 2: Subtree Selection for Problem Integration
Based on your role analysis, select a subtree (tree-structured subset) where all selected leaf features can be naturally integrated into a single, high-quality competitive programming problem.
- Only include features that contribute meaningfully to the same problem idea.
- Internal nodes are included only if they have selected children.
- For each selected leaf, include only its "feature" name and "potential_use".
Step 3: Integration Strategy
Briefly describe ("integration_strategy") how the selected features can be integrated together in a single problem, focusing on how their combination enables a meaningful and challenging algorithmic scenario.

Return a JSON object **with exactly this structure** (an example):
{{
  "feature_roles_tree": {{
    "algorithm": {{
      "search algorithm": {{
        "binary search": {{
          "recursive binary search": {{
            "potential_use": "Used for divide-and-conquer searching in sorted structures or answer spaces."
          }},
          ...
  }},
  
  "selected_features_tree": {{
    "algorithm": {{
      "search algorithm": {{
        "binary search": {{
          "recursive binary search": {{
            "feature": "recursive binary search",
            "potential_use": "Used for divide-and-conquer searching in sorted structures or answer spaces."
          }}
    ...
  }},

  "integration_strategy": "The problem will require recursive binary search to efficiently search over a sorted value space, while bitwise AND operations will be used to filter candidate solutions according to constraints. Their combination allows for a problem that involves searching over sets and optimizing bitwise criteria."
}}

**Available Features (Tree):**
{features_json}
\end{appendixjsonlisting}

\newpage
\subsubsection{Compatibale Feature Selection}
\label{appendix_sec:Compatibale Feature Selection}
We present a case to examine how model selects compatible features and combine them.

Given a sampled feature tree:
\begin{appendixjsonlisting}
"input_features": {
    "algorithms": {
      "graph_algorithms": {
        "shortest_path": [
          "Dijkstra's algorithm",
          "Floyd-Warshall"
        ],
        "network_flow": [
          "Ford-Fulkerson",
          "Edmonds-Karp"
        ]
      },
      ...
\end{appendixjsonlisting}

LLM pairs each feature with \textit{potentially usage} to obtain feature tree with role annotation. For example, LLM will anonote feature ``rolling hash" as ``Compute hash values for sliding windows in constant time". These annotations help LLM to aggregate these features based on their potentially usage.
For above given feature tree, the feature tree with potential usage looks like:
\begin{appendixjsonlisting}
"feature_roles_tree": {
    "algorithms": {
      "graph_algorithms": {
        "shortest_path": {
          "Dijkstra's_algorithm": {
            "potential_use": "Find single-source shortest paths in weighted graphs with non-negative edges"
          },
          "Floyd_Warshall": {
            "potential_use": "Compute all-pairs shortest paths with O(n^3) complexity"
          }
        },
        ...
\end{appendixjsonlisting}

LLM then selects a compatible and consistent subtree that can formulate a self-contained competitive programming problem. LLM keeps the features that can be aggregated into selected feature tree, and concluding with an integration strategy, which displays how to combine these features into a unified problem. In this case, LLM selects "Dijkstra's algorithm", "Edmonds-Karp", "segment tree", and "tree DP", and aims to formulate a problem around "dynamic network optimization".
\newpage
\begin{appendixjsonlisting}
"algorithms": {
  "graph_algorithms": {
    "shortest_path": {
      "Dijkstra's_algorithm": {
        "feature": "Dijkstra's algorithm",
        "potential_use": "Primary pathfinding algorithm"
      }
    },
    "network_flow": {
      "Edmonds_Karp": {
        "feature": "Edmonds-Karp",
        "potential_use": "Flow computation with guaranteed complexity"
      }
    }
  }
},
"data_structures": {
  "tree_structures": {
    "segment_tree": {
      "feature": "segment tree",
      "potential_use": "Maintain dynamic edge weights or capacities"
    }
  }
},
...
"integration_strategy": "Create a dynamic network optimization problem where Dijkstra's algorithm finds shortest paths that are used as augmenting paths in a modified Edmonds-Karp flow algorithm. Use segment tree to handle dynamic updates to edge capacities based on flow history. Apply tree DP on the shortest path tree to compute optimal flow distributions. This models a transportation network with time-varying capacities."
\end{appendixjsonlisting}

\subsubsection{From Feature to Stylized Task}
\label{appendix_sec:From Feature to Task Generation}
We separate feature selection from task generation, as our initial attempts showed that prompting an LLM to perform both within a single prompt often led it to choose fewer features and produce overly simple problems.

\begin{figure*}[t]
    \centering
    {\color[rgb]{0.00,0.45,0.70}\fbox{\includegraphics[width=0.82\textwidth]{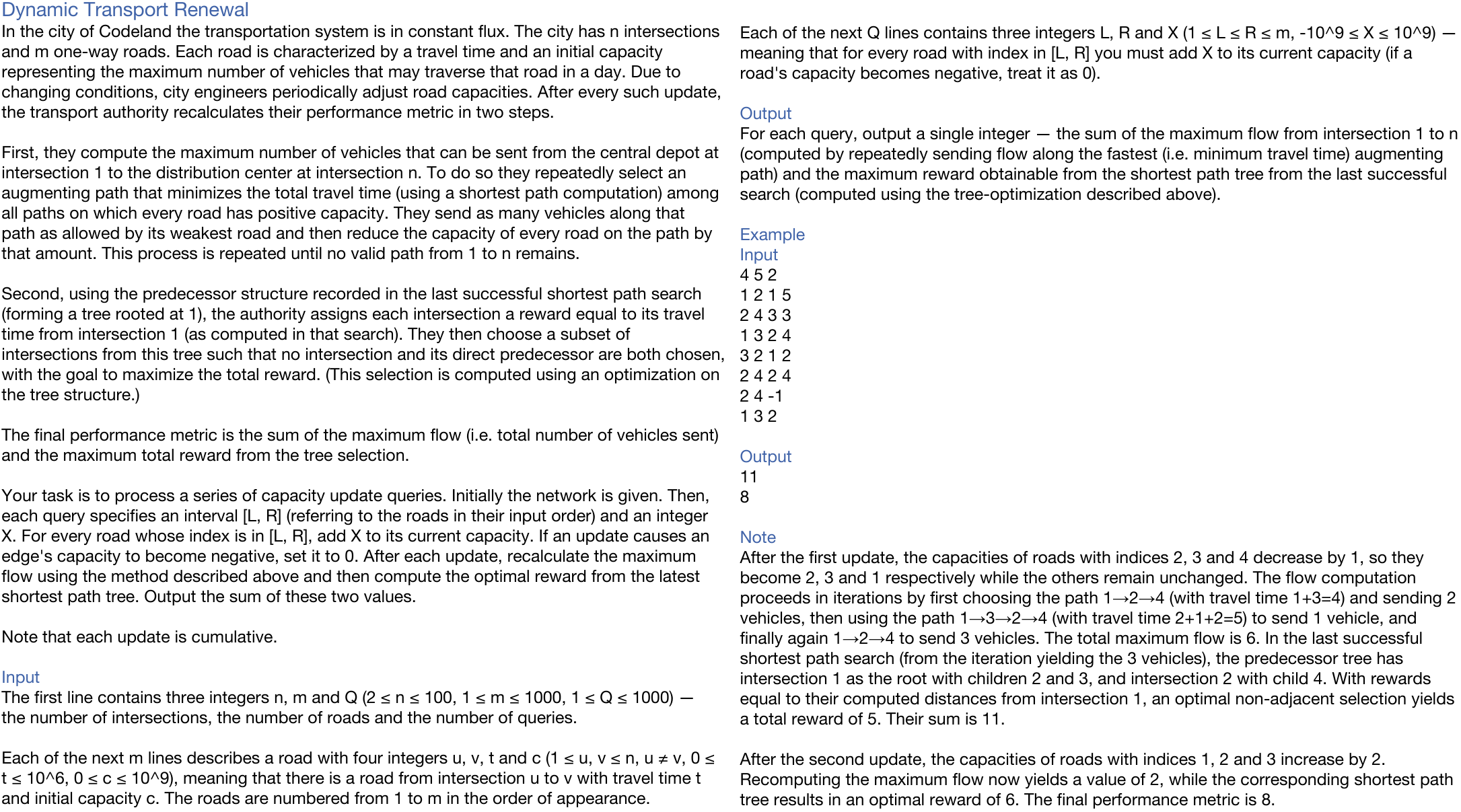}}}
    \caption{Case for Codeforces-style Problem, featuring rich, imaginable narrative contexts.}
    \label{fig:codeforces}
\end{figure*}

During task generation, LLM receives \textit{selected features tree} and its \textit{integration strategy} to formulate stylized task based on prompt recieved. In this instance, our generated Codeforces problem is shown in Figure~\ref{fig:codeforces}.

\begin{figure}[!b]
    \centering
    \includegraphics[width=0.80\columnwidth]{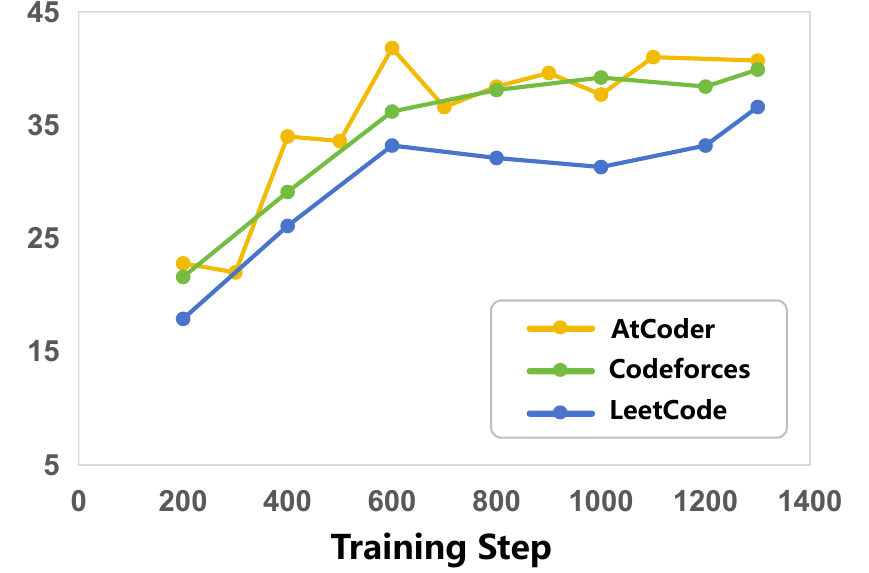}
    \caption{Task style comparison.}
    \label{fig:style_comparison_appendix}
\end{figure}

\setlength{\fboxsep}{3pt}   % 图片和边框的间距
\setlength{\fboxrule}{0.5pt}  % 边框粗细

The rationale for above two-stage approach is that a single-step approach is less effective. When performing both steps simultaneously, LLMs tend to oversimplify complex instructions into trivial cases, reducing both diversity and difficulty of the generated task.

To empirically validate this, we generated 32k tasks using the one-step method (feature-tree $\rightarrow$ task) and using proposed ``two-stage'' method (feature-tree $\rightarrow$ sub-tree $\rightarrow$ task). The SFT results on LiveCodeBench v5 are as Table~\ref{single_vs_two_step}. The 5.3 gain shows that explicit sub-tree selection and integration is significantly helpful for producing high-quality, challenging tasks and justifies the strategy's modular design.

\begin{table}[!t]
\centering
\small
\setlength{\tabcolsep}{5pt}
\caption{One-step and two-stage generation.}
\label{single_vs_two_step}
\begin{tabular}{@{}lc@{}}
\toprule
\textbf{Generation Method} & \textbf{V5 Score (avg@4)} \\
\midrule
One-Step (end-to-end) & 34.8 \\
Two-Stage & 40.1 \\
\bottomrule
\end{tabular}
\end{table}

\subsubsection{Ablation on Task Style}
\label{appendix_sec: Ablation on Task Style}
We evaluate the effect of task styles (AtCoder, Codeforces, and LeetCode) by synthesizing three corpora of 32k tasks each (8k unique problems with 4 solutions per problem) from identical input features. For each corpus, solutions are generated with DeepSeek-R1-0528 and used to fine-tune the Qwen2.5-Coder-7B-Instruct. Results are shown in Figure~\ref{fig:style_comparison_appendix}. Although AtCoder-style tasks yield slightly higher scores, we adopt Codeforces-style as the predominant format in our demonstration dataset (Codeforces : AtCoder : LeetCode = 70 : 15 : 15), reflecting its prominence as the mainstream competitive-programming platform.

\subsection{Task Difficulty Estimates}
\label{appendix_sec: Task Difficulty Estimates}

Judging the difficulty of a synthetic task is challenging. To better capture the difficulty distribution of tasks generated by \method, we adopt a classifier-based approach. 
Specifically, we add a special classification token to Qwen2.5-Coder-14B-Instruct and fine-tune it to predict the Codeforces rating of 6,246 tasks from the CodeContests dataset with annotated ratings, reserving 5\% as a validation set. 
The fine-tuned model achieves 84\% classification accuracy on the validation set. 
We then use this model to estimate the difficulty of 1,000 tasks generated by the proposed method, obtaining a holistic distribution as shown in Table~\ref{tab:difficulty_dist}. 

\begin{table}[!h]
  \centering
  \scriptsize
  \caption{Difficulty distribution of Codeforces-style ratings.
  ``Original'' denotes the annotated distribution from CodeContests,
  and ``Test'' denotes 1,000 tasks generated by the proposed method.}
  \label{tab:difficulty_dist}
  \resizebox{\columnwidth}{!}{%
  \begin{tabular}{@{}crrrr@{}}
    \toprule
    \textbf{CF Rating} & \textbf{Original} & \textbf{Test (Ours)} & \textbf{Original Share} & \textbf{Test Share} \\
    \midrule
    1200 & 623 & 0   & 10.0\% & 0.0\% \\
    1400 & 727 & 0   & 11.7\% & 0.0\% \\
    1600 & 889 & 0   & 14.3\% & 0.0\% \\
    1800 & 840 & 16  & 13.5\% & 1.6\% \\
    2000 & 797 & 2   & 12.8\% & 0.2\% \\
    2200 & 697 & 47  & 11.2\% & 4.7\% \\
    2400 & 665 & 585 & 10.7\% & 58.5\% \\
    2600 & 484 & 319 & 7.8\%  & 31.9\% \\
    2800 & 312 & 12  & 5.0\%  & 1.2\% \\
    3000 & 233 & 15  & 3.7\%  & 1.5\% \\
    3200 & 157 & 4   & 2.5\%  & 0.4\% \\
    3400 & 122 & 0   & 2.0\%  & 0.0\% \\
    \midrule
    \textbf{Total} & \textbf{6,246} & \textbf{1,000} & \textbf{100\%} & \textbf{100\%} \\
    \bottomrule
  \end{tabular}%
  }
\end{table}

The estimated ratings are concentrated at 2400--2600 (90.4\%), reflecting our deliberate focus on challenging but learnable problems and likely contributing to weaker gains on the easy split. This should not be conflated with problem diversity: our diversity claim concerns coverage across algorithms, data structures, problem types, and implementation patterns, rather than a balanced rating histogram. Moreover, the 84\%-accurate classifier is only a coarse estimator, so we do not treat these predicted ratings as ground truth.

\subsection{Task Diversity Estimates}
To analyze the diversity of our generated tasks quantitatively, we analyze diversity in the embedding space following the steps below:
(i) Embedding: We first embed the tasks into embeddings using \textit{jinaai/jina-embeddings-v2-base-code}, a specialized coding embedding model.
(ii) t-SNE Dimensionality Reduction: We apply t-SNE to reduce the embedded data to 2D space.
(iii) Clustering: We perform K-means clustering on the t-SNE-reduced data to group the data into 10 clusters and compute the centroids of each cluster.
(iv) Inter-cluster Distance Calculation: We calculate the Euclidean distance between cluster centroids. Larger inter-cluster distances indicate greater diversity within the dataset.

\begin{figure*}[t]
  \centering 
  
  \begin{subfigure}[t]{0.44\textwidth}
    \centering
    \includegraphics[width=0.78\linewidth]{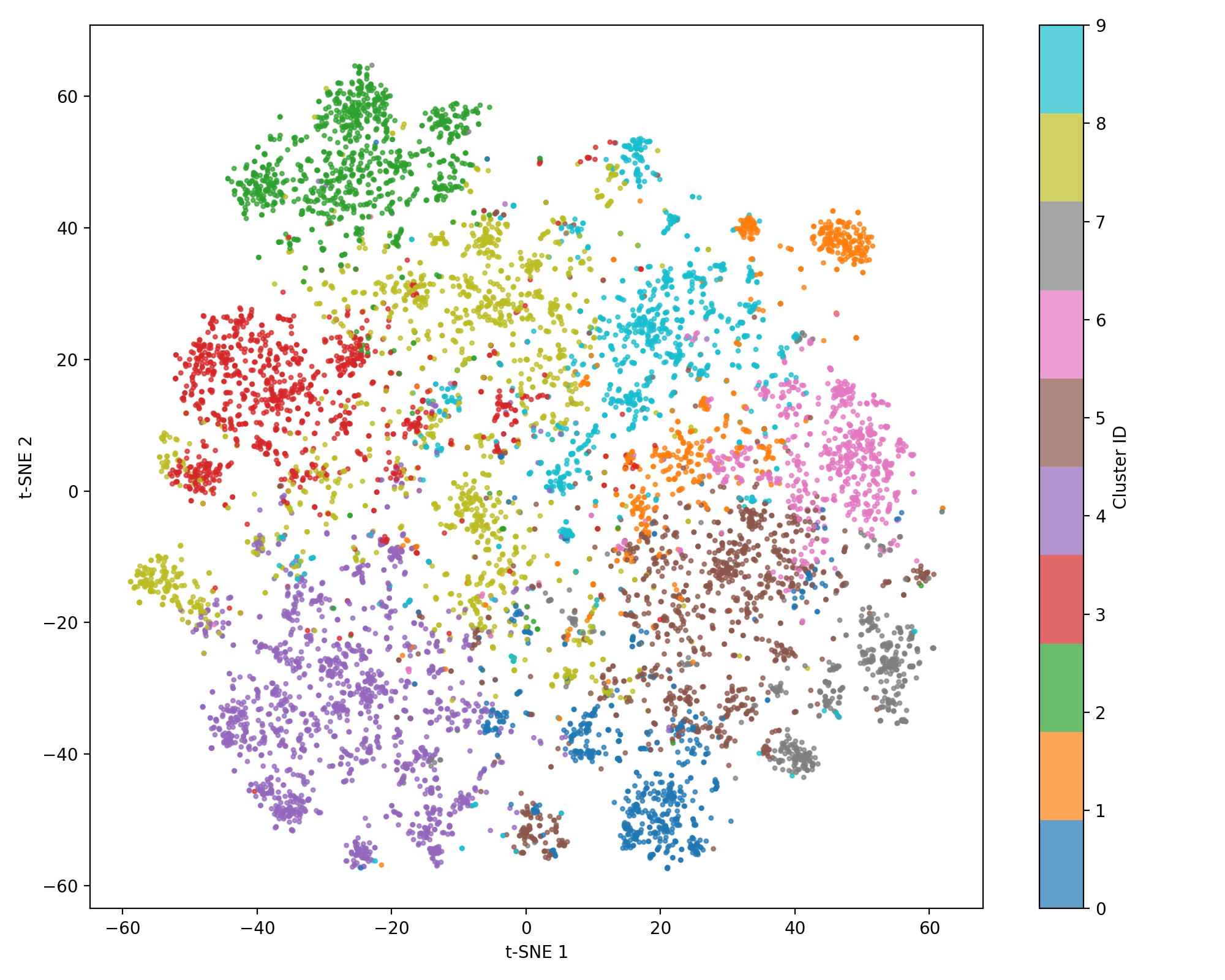}
    \caption{t-SNE visualization of our datasets.}
    \label{fig:tsne1}
  \end{subfigure}
  \hfill
  \begin{subfigure}[t]{0.44\textwidth}
    \centering
    \includegraphics[width=0.78\linewidth]{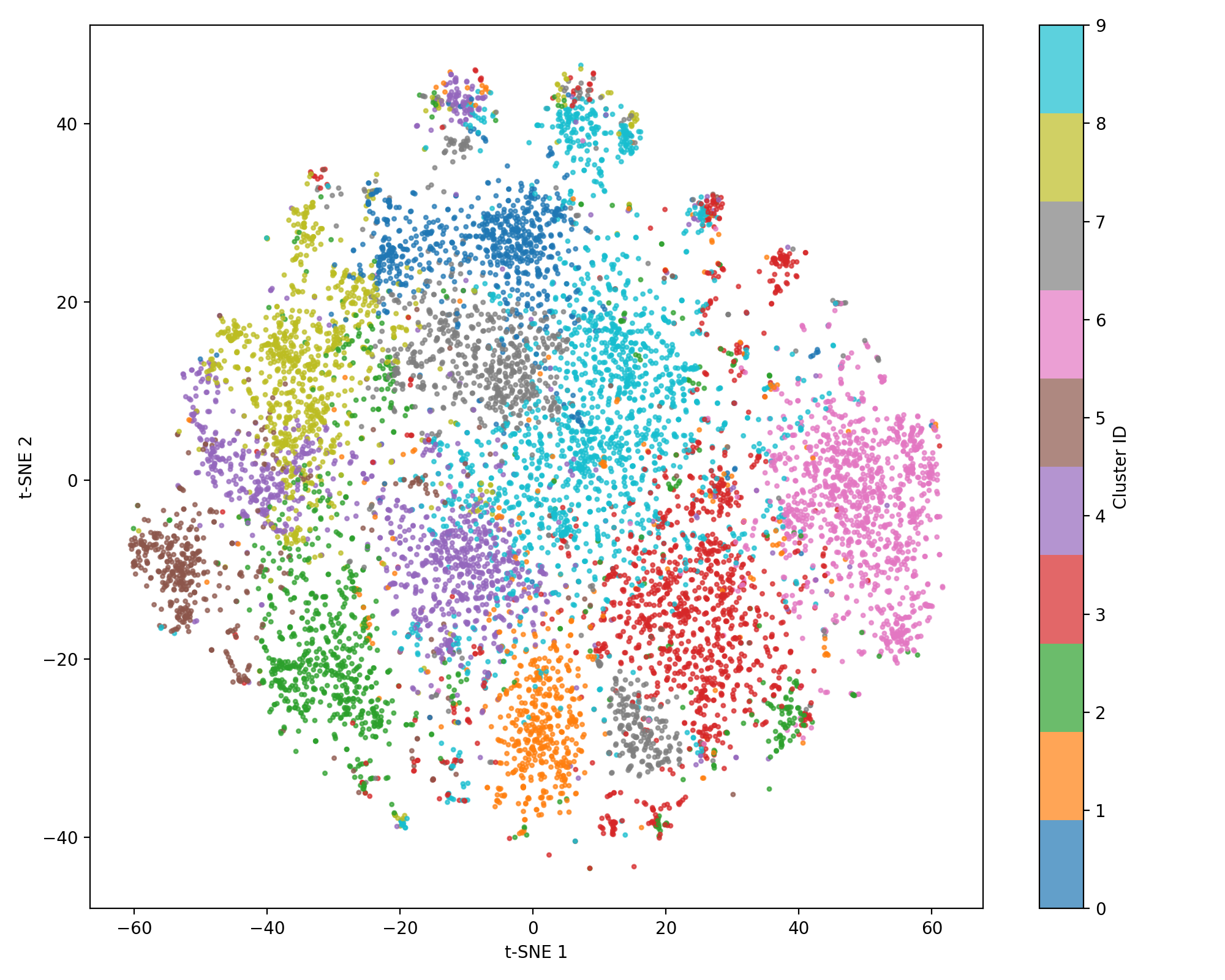}
    \caption{t-SNE of the Evol-Instruct-Code.}
    \label{fig:tsne2}
  \end{subfigure}
  \caption{Task diversity visualizations in the t-SNE space.}
  \label{fig:tsne_appendix}
  
\end{figure*}

In our datasets (randomly sampled 10k), cluster sizes range 529-1,612 items, average centroid distance 0.613, min 0.369, max 0.760. In Evol-Instruct-Code, the mean centroid distance is 0.507.
The visualization results are shown in Figure~\ref{fig:tsne1} and Figure~\ref{fig:tsne2}. The visualization suggests that the clusters in our dataset are more widely separated compared to those in Evol-Instruct-Code, indicating higher diversity.

\subsection{Comparison with Alternative Synthesis Strategies}
\label{appendix_sec:synthesis_comparison}

As presented in Table~\ref{tab:selfcodealign_vs_synthsmith}, we compare the proposed strategy against two baselines to validate our methodological choices. First, compared to EpiCoder (using general-purpose prompts), the domain-adapted synthesis achieves a substantial 20.9\% improvement. Since this method is built upon EpiCoder's framework but injects competitive programming-oriented feature extraction and evolution, this gain directly validates the necessity of domain expertise for synthesizing high-complexity tasks. General-purpose synthesis is insufficient for expert-level domains. Second, compared to SelfCodeAlign~\citep{wei2024selfcodealign} (where we applied the same domain adaptation), the proposed method still yields a 4.6\% gain. This suggests that for competitive programming, the feature-based evolution strategy is more effective than concept composition at generating high-quality, diverse, and rigorous training data.

\begin{table}[!ht]
\centering
\small
\caption{Synthetic Data Comparison.}
\label{tab:selfcodealign_vs_synthsmith}
\begin{tabular}{lcccc}
\toprule
\textbf{Method} & \textbf{Evolution} & \textbf{Adapted}  & \textbf{LCB v5} \\
\midrule
SelfCodeAlign-10k & \xmark & \cmark& 27.1 \\
Ours-10k & \cmark & \cmark & \textbf{31.7}  \\
EpiCoder-64k & \cmark & \xmark  & 25.4 \\
Ours-64k & \cmark & \cmark  & \textbf{46.3}  \\
\bottomrule
\end{tabular}
\end{table}

\newpage
\section{Solution Generation and Quality Assurance}
\subsection{Validation on Solution}
\label{appendix_sec: Validation on Solution}
For tasks with descriptions shorter than 200 tokens, we discard them, as such descriptions are often either too trivial or incomplete. For each generated solution, we ensure quality by (i) removing samples without complete think and answer tags, (ii) rejecting cases where the extracted Python block fails AST validation, (iii) excluding solutions that contain multiple code blocks after the reasoning process, as they hinder reliable solution extraction, and (iv) filtering out samples exceeding 25k tokens to prevent overthinking and to reduce SFT cost caused by sequence padding.

\subsection{SFT Dataset Statistics}
\label{appendix_sec: SFT Dataset Statistics}
\begin{figure}[!h]
    \centering
    \includegraphics[width=0.88\columnwidth]{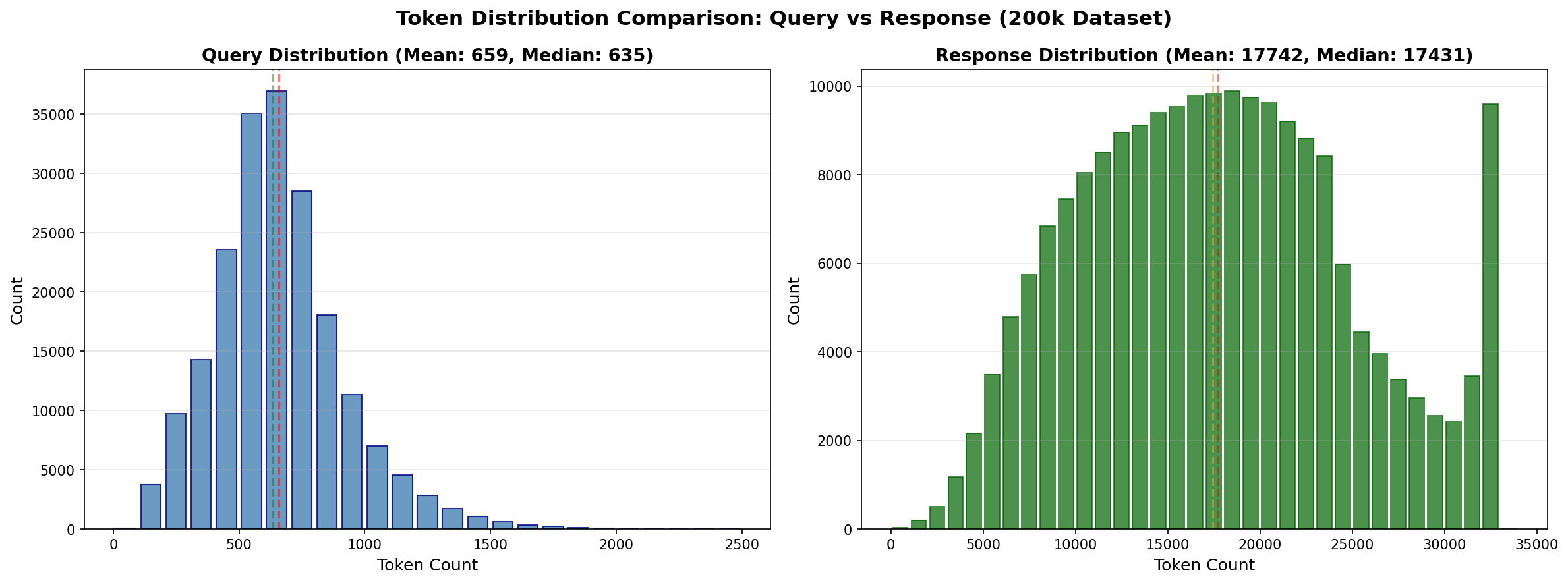}
    \caption{Dataset statistics of the demonstration dataset.}
    \label{fig:dataset_statistics}
\end{figure}

The overall token length distribution, shown in Table~\ref{tab:dataset_statistics}, and Figure~\ref{fig:dataset_statistics}, primarily follows a normal distribution, with a median of 16k.

\begin{table*}[!t]
\centering
\small
\caption{Token statistics for tasks and solutions of the demonstration dataset.}
\label{tab:dataset_statistics}
\begin{tabular}{@{}lcccccc@{}}
\toprule
\textbf{Type} & \textbf{Min} & \textbf{Max} & \textbf{Mean} & \textbf{Median} & \textbf{Std Dev} & \textbf{Total Tokens} \\
\midrule
Task     & 200    & 3,537  & 658.91   & 635.00   & 258.49   & 134.3M \\
Solution & 1,711  & 33,144 & 17,742.50 & 17,431.00 & 7,295.92 & 3.25B \\
\midrule
Dataset Size & \multicolumn{5}{c}{200,091 entries}  & 3.38B \\
\bottomrule
\end{tabular}%
\end{table*}

\newpage
\section{Test Case Generation}
\label{appendix_sec: test case generation}
\subsection{Prompting-based Test Generation}
\begin{appendixjsonlisting}
You are a professional test case generation expert, skilled at designing comprehensive test cases for programming problems. Please generate 15 different test cases for the following programming problem, including edge cases, small-scale, medium-scale, and large-scale test data.

Problem:
{problem_statement}

Requirements:
1. Generate 15 test cases
2. Include edge cases (empty input, minimum values, maximum values, etc.)
3. Include different scales of data (small, medium, large)
4. Each test case should have clear input data
5. Ensure test cases can thoroughly validate the correctness of solutions

Please return in JSON format as follows:
{{
    "test_cases": [
        {{
            "idx": 0,
            "description": "Test case description",
            "input_string": "Input data"
        }},
        ...
    ]
}}
\end{appendixjsonlisting}

\begin{table*}[!h]
\centering
\small
\caption{Comparison of Prompting-based and Tool-based Test Generation. The tool-based approach excels in test diversity, accuracy, and the ability to generate more challenging test cases.}
\label{tab:ablation_test_generation}
\resizebox{\textwidth}{!}{%
\begin{tabular}{@{}lcccccccccc@{}}
\toprule
 & \textbf{Random} & \textbf{Scalable} & \textbf{Boundary} & \textbf{Stress} & \textbf{Cost} & \textbf{Avg Tests} & \textbf{Min Tests} & \textbf{Max Tests} & \textbf{Consensus} & \textbf{Pass Rate} \\
\midrule
Prompting-based & \xmark & \xmark & \xmark & \xmark & low & 13.6 & 5 & 15 & 82.0\% & 77.4\% \\
Tool-based & \checkmark & \checkmark & \checkmark & \checkmark & high & 18.3 & 5 & 27 & 78.8\% & 87.9\% \\
\bottomrule
\end{tabular}%
}
\end{table*}

\subsection{Comparison of Test Generation Methods}
\label{appendix_sec: test case comparison}

This comparison is scoped to our synthesis pipeline and is not intended as a comprehensive benchmark of test-case generation methods. Recent work has proposed stronger test synthesis, feedback-driven refinement, grammar-guided generation, targeted generator generation, and adversarial hack-style tests. These techniques are orthogonal to our pipeline and could be incorporated to improve the coverage and discriminative power of the generated rewards.
\citep{he2025hardtests,wang-etal-2025-codecontests,cai2026codecontestsopoweringllmsfeedbackdriven,ijcai2025p861,cao2026llmsgeneratereliabletest,shi2026codehackerautomatedtestcase}

We compare prompting-based and tool-based test generation using tasks from CodeContests~\citep{li2022competition}. We leverage the corresponding golden solutions to evaluate the accuracy and complexity of the tests produced by the two approaches. The results in Table~\ref{tab:ablation_test_generation} show that the tool-based approach outperforms the prompting-based method across multiple dimensions. Qualitatively, it is more versatile, capable of systematically generating random, scalable, boundary, and stress tests, which are essential for robust code evaluation but not supported by prompting-based methods.

Quantitatively, the tool-based approach achieves a higher pass rate on ground-truth solutions (87.9\% vs. 77.4\%), confirming that its test cases are more accurate and reliable. It also generates more challenging and discriminative tests, as reflected by the lower consensus ratio (78.8\% vs. 82.0\%), which indicates stronger effectiveness in uncovering subtle bugs. In addition, the tool-based generator provides broader test coverage, albeit at a higher computational cost.

\subsection{Tool-based Test Generation}

The tool-based test generation strategy relies on \textbf{CYaRon}, an open-source Python library aimed at rapidly generating random data for Informatics Olympiad problems (or problems of equivalent difficulty). This library contains a variety of common data structures (e.g., graphs, trees, polygons, vectors, strings, and sequences), along with mathematics-related functions and the necessary input/output interfaces. When prompting the Teacher model to utilize the CYaRon tool, we provide its detailed documentation and usage instructions as part of the prompt. Additionally, we encourage the model to generate more boundary tests and large-scale random use cases. To ensure the sufficiency of test cases, we mandate the use of this library in conjunction with its random features and set a seed to ensure reproducibility. The detailed prompt used is illustrated as:

\begin{appendixjsonlisting}
Please write a test case generator that meets the following requirements based on the following CYaRon documentation:

1. Write a canonical CYaRon Generator using Python
2. Generate a single, executable Python program that can produce test cases with at least 5 different features
3. The Python program should save each test case individually in the format [use case characteristics].in
4. The program should include a variety of test case types such as base cases, boundary cases, large random cases, etc
5. The Python program code should contain clear comments to explain the design intent for each test case generation
6. The .in output files should contain ONLY pure input data without any comments, explanations, or answer validation
7. The Python program should be able to generate all test cases in a single run when executed
8. The program should use argparse to provide configurable random seed control:
   parser.add_argument('--seed', type=int, default=42, help='Random seed for reproducibility')
9. All random number generation must use Python's built-in random module (import random) - do not use any external random libraries or the random functions from CYaRon

### CYaRon Documentation
[The complete API documentation of the CYaRon library is provided here, covering Input/Output (IO), Graph Generation, Polygon, Vector, String, Sequence, and Utilities. We omit the full text for brevity as it follows the standard documentation of the library.]

### Code Question
{QUESTION}
\end{appendixjsonlisting}

\newpage
\section{Dual-verification}
\label{appendix_sec: Dual-verification}

\subsection{Core Algorithm}
\label{appendix_sec:dual-verification-algorithm}

We summarize the symbols used in the dual-verification process in
Table~\ref{tab:notation}, and outline the corresponding procedure in
Algorithm~\ref{alg:dual_verification}.

\begin{table}[!h]
\centering
\small
\caption{Notation for Our Framework.}
\label{tab:notation}
\begin{tabular}{@{}p{0.30\columnwidth}p{0.62\columnwidth}@{}}
\toprule
$\{x_i\}_{i=1}^{n}$       & Test inputs for a task $q$ \\
$\{A^j\}_{j=1}^{m}$       & Candidate solutions (LLM-generated) \\
$y_i^j$                   & Output of $A^j$ on input $x_i$ \\
$\hat{y}_i$               & Provisional label via majority vote on $\{y_i^j\}_{j=1}^m$ \\
$w_i$                     & Difficulty weight for $x_i$ \\
$\mathcal{T}_{candidate}$ & Provisional labeled set $\{(x_i,\hat{y}_i,w_i)\}$ \\
$\mathcal{T}_{golden}$    & Weighted suite for selecting the solution \\
$\mathcal{T}_{val}$       & Hold-out validation set \\
$S_j$                     & Weighted score of $A^j$ on $\mathcal{T}_{golden}$ \\
$A_{golden}$              & Final selected ``golden'' solution \\
\bottomrule
\end{tabular}
\end{table}

\begin{algorithm}[tb]
   \caption{Dual-Verification of Solutions and Test Cases (Strict Verification)}
   \label{alg:dual_verification}
\begin{algorithmic}[1]
   \REQUIRE Task \(q\); test inputs \(\{x_i\}_{i=1}^n\); candidate solutions \(\{A^j\}_{j=1}^m\).
   \ENSURE Golden solution \(A_{\mathrm{golden}}\) and test suite \(\mathcal{T}_{\mathrm{golden}}\), or \textbf{None}.
   
   \STATE \textbf{Step 1: Consensus Voting \& Weighting}
   \FOR{$i = 1$ to $n$}
      \FOR{$j = 1$ to $m$}
         \STATE Run $y_i^j \gets A^j(x_i)$
      \ENDFOR
      \STATE $\hat{y}_i \gets \arg\max_{y}\sum_{j=1}^m \mathbb{I}(y_i^j=y)$
      \STATE $w_i \gets \mathrm{Weight}(x_i)$
   \ENDFOR
   \STATE $\mathcal{T}_{\mathrm{candidate}} \gets \{(x_i,\hat{y}_i,w_i)\}_{i=1}^n$
   
   \STATE \textbf{Step 2: Split Candidate Set}
   \STATE Randomly partition $\mathcal{T}_{\mathrm{candidate}}$ into $\mathcal{T}_{\mathrm{golden}}$ and $\mathcal{T}_{\mathrm{val}}$
   
   \STATE \textbf{Step 3: Weighted Selection}
   \FOR{$j = 1$ to $m$}
      \STATE $S_j \gets \sum_{(x_i,\hat{y}_i,w_i)\in \mathcal{T}_{\mathrm{golden}}} w_i \cdot \mathbb{I}(A^j(x_i)=\hat{y}_i)$
   \ENDFOR
   \STATE $j^\star \gets \arg\max_j S_j$
   \STATE $A'_{\mathrm{golden}} \gets A^{j^\star}$
   
   \STATE \textbf{Step 4: Hold-out Confirmation}
   \STATE Compute unweighted accuracies of all $A^j$ on $\mathcal{T}_{\mathrm{val}}$
   \STATE $j^{\dagger} \gets \arg\max_j \text{Acc}(A^j,\mathcal{T}_{\mathrm{val}})$
   
   \IF{$j^{\dagger}=j^\star$}
      \STATE $A_{\mathrm{golden}} \gets A'_{\mathrm{golden}}$
      \STATE \textbf{return} $A_{\mathrm{golden}}, \mathcal{T}_{\mathrm{golden}}$
   \ELSE
      \STATE \textbf{return} \textbf{None} \COMMENT{Discard task}
   \ENDIF
\end{algorithmic}
\end{algorithm}

\subsection{Test-Case Weighting Criteria}
\label{appendix_sec:Test-Case Weighting Criteria}

We offer two distinct strategies for assigning weights to individual test cases:

\textbf{Semantic-Based Weighting.} During test-case generation, the model is prompted to produce multiple categories of test cases (stored as \texttt{.in} files), including nominal (weight = 1), complex (2), boundary (3), and stress (4) scenarios. This assigns higher weights to test cases that are more likely to expose corner cases or failure modes.

\textbf{Size-Based Weighting.} We assign weights based on the size of the input files, which serves as a proxy for memory consumption. Specifically, we sort test cases by the size of their input files and divide them into four equal-sized buckets: the smallest 25\% receive weight = 1, the next 25\% receive weight = 2, the next 25\% receive weight = 3, and the largest 25\% receive weight = 4. This ensures that heavier test cases, which require greater memory resources, are assigned higher weights.

In this work, we adopted Size-Based Weighting for data synthesis.

\subsection{Validation Effectiveness Analysis}
\label{appendix_sec:validation-effectiveness}

\subsubsection{Error Rate for Labeling Test Outputs via Voting}
\label{appendix_sec:Error Rate for Labeling Test Outputs via Voting}

On TACO-verified, we measure a 5.27\% false-positive rate under voting with 8 solutions. To assess the false-positive rate of test-output labeling, we evaluate the proposed approach on real-world, verified datasets. Specifically, we randomly sample 500 tasks from the TACO-verified dataset, and for each task, we randomly retain 20 test cases.

For each task, we generate $n$ ($n \in \{4, 8, 16\}$) candidate solutions using R1-0528, perform majority voting on the outputs for each test input, and compare the voted consensus output against the ground-truth output to obtain a quantitative labeling accuracy. The resulting test-output labeling accuracy under different values of $n$ is shown in Table~\ref{tab:labeling_accuracy} and Table~\ref{tab:source_breakdown}.

\begin{table}[h]
    \centering
    \small
    \caption{Average Test Output Labeling Accuracy with varying $n$.}
    \label{tab:labeling_accuracy}
    \begin{tabular}{@{}cc@{}}
        \toprule
        \textbf{$n$ (\# solutions)} & \textbf{Labeling Accuracy} \\
        \midrule
        $4$  & $94.39\%$ \\
        $8$  & $94.73\%$ \\
        $16$ & $95.13\%$ \\
        \bottomrule
    \end{tabular}
\end{table}

\begin{table}[h]
    \centering
    \small
    \caption{Test Output Labeling Accuracy across different sources.}
    \label{tab:source_breakdown}
    \begin{tabular}{@{}lccc@{}}
        \toprule
        \textbf{Source} & \textbf{$n=4$} & \textbf{$n=8$} & \textbf{$n=16$} \\
        \midrule
        AtCoder    & 94.75\% & 95.00\% & 96.61\% \\
        CodeChef   & 92.80\% & 92.80\% & 92.80\% \\
        CodeForces & 94.44\% & 94.81\% & 95.06\% \\
        \bottomrule
    \end{tabular}
\end{table}

Increasing the number of sampled solutions consistently improves test output labeling accuracy. With $n = 8$, the false-positive rate is 5.27\%, which falls within an acceptable range and demonstrates that the approach is potentially reliable to be transferred to the synthetic setting.

\subsubsection{Same-Family vs. Cross-Family Voting}
To test correlated same-source errors, we hold 500 TACO-verified tasks and their test inputs fixed while changing only the candidate-solution pool. At a matched pool size of four, combining two R1-0528 and two Qwen3-235B solutions improves labeling accuracy by 2.64 points over four R1-0528 samples. Increasing the heterogeneous pool to eight provides a smaller additional gain.

\begin{table}[!h]
\centering
\small
\caption{Test-output labeling accuracy by voting pool.}
\label{tab:cross_family_voting}
\begin{tabular}{@{}lc@{}}
\toprule
\textbf{Voting Pool} & \textbf{Accuracy} \\
\midrule
R1-0528 $\times 4$ & 89.91\% \\
R1-0528 $\times 2$ + Qwen3-235B $\times 2$ & 92.55\% \\
R1-0528 $\times 4$ + Qwen3-235B $\times 4$ & \textbf{92.94\%} \\
\bottomrule
\end{tabular}
\end{table}

Cross-family voting reduces, but cannot rule out, correlated teacher errors; the human audit below measures the remaining error directly on synthetic tasks.

\subsection{Solution Quality Assessment}
\label{appendix_sec:solution-quality-assessment}

\subsubsection{Error Rate of Golden Solution}
\label{appendix_sec:Error Rate of Golden Solution}

To enable quantitative assessment, we adopt two evaluations: (1) measuring the error rate of dual verification on our synthetic datasets, which yields pass rate distributions across various proprietary LLMs; and (2) evaluating the actual error rate on real-world datasets (TACO-verified), resulting in a 7.85\% error rate.

\paragraph{(i) Synthetic Task Evaluation.}
We first use DeepSeek-R1-0528 to generate multiple candidate solutions for each synthetic task. We then apply our dual-verification strategy to select the golden solution and measure its pass rates on the voted test cases. The pass rate distribution is shown in Table~\ref{tab:golden_pass_rates}.

\begin{table}[h]
    \centering
    \small
    \caption{Distribution of Golden Solution Pass Rates on Voted Test Cases using R1-0528.}
    \label{tab:golden_pass_rates}
    \setlength{\tabcolsep}{12pt}
    \begin{tabular}{@{}lc@{}}
        \toprule
        \textbf{Range (\%)} & \textbf{Ratio (\%)} \\
        \midrule
        $(0, 20)$   & $13.12$ \\
        $[20, 40)$  & $17.29$ \\
        $[40, 60)$  & $17.57$ \\
        $[60, 80)$  & $14.94$ \\
        $[80, 100)$ & $13.39$ \\
       
        $100$       & $\mathbf{23.66}$ \\ 
        \bottomrule
    \end{tabular}
\end{table}

Here, each percentage range represents the fraction of tasks whose selected golden solution attains a pass rate within that interval. For example, the $[80, 100)$ range indicates that 13.39\% of tasks have golden solutions that pass between 80\% and 100\% of their voted test cases, while 23.66\% of the solutions pass all test cases.

Note that solution quality is strongly tied to model capability. The pass rates of the proprietary models are presented in Table~\ref{tab:prop_llm_pass_distribution} in the task solvability analysis section.

If we adopt a more capable model such as GPT-5-High, 66.98\% of the tasks can be solved perfectly in a single attempt.

\paragraph{(ii) Real-world Dataset Evaluation.}
We also apply our dual-verification approach to real-world, verified datasets to measure the error rate of the selected golden solutions. Because real-world datasets contain ground-truth test cases, the resulting error rate accurately reflects the true quality of the selected solutions.

Specifically, we randomly select 500 tasks from the TACO-verified dataset, each with 20 retained test cases as ground truth tests. We apply our dual-verification procedure using R1-0528 to label test outputs via voting, and then select the golden solution based on the pass rate on the voted test cases. We then evaluate each golden solution against the ground-truth tests.

The verification results under different numbers of candidate solutions ($n$) are shown in Table~\ref{tab:taco_verification}.

\begin{table}[h]
    \centering
    \small
    \caption{Verification results on TACO-verified dataset with varying candidate solutions ($n$).}
    \label{tab:taco_verification}
    % \resizebox{\columnwidth}{!}{%
    \begin{tabular}{@{}ccc@{}}
        \toprule
        \textbf{$n$} & \textbf{Avg. Pass Rate} & \textbf{Full Pass Rate} \\
        (Candidates) & (test-case level) & (task-level) \\
        \midrule
        4  & 91.79\% & 84.20\% \\
        8  & 92.15\% & 85.00\% \\
        16 & 92.50\% & 85.80\% \\
        \bottomrule
    \end{tabular}%
    % }
\end{table}

On the TACO-verified dataset, the proposed approach yields a 7.85\% error rate in the selected golden solutions when $n=8$. The error rate further decreases as the number of rollout solutions increases. Such an error level is acceptable, indicating that the approach has the potential to be transferred to the synthetic setting.

\subsubsection{Human Audit of the Synthetic Set}
\label{appendix_sec:human-audit}
We randomly sample 150 tasks from the SFT corpus and manually assess three binary criteria: whether each task is well-defined, whether its selected golden solution is correct, and whether the voted outputs of its generated tests are correct. This audit directly evaluates the synthetic set instead of relying on TACO ground truth.

\begin{table}[!h]
\centering
\small
\caption{Human audit of 150 synthetic SFT tasks.}
\label{tab:human_audit}
\begin{tabular}{@{}lc@{}}
\toprule
\textbf{Criterion} & \textbf{Rate} \\
\midrule
Well-defined task & 90.0\% \\
Correct golden solution & 87.3\% \\
Correct voted test labels & 87.3\% \\
\bottomrule
\end{tabular}
\end{table}

The audit estimates a 12.7\% golden-solution error rate and the same residual rate for voted labels. Among incorrect labels, 94\% arise from one mechanically detectable failure mode: function-signature tasks lacking a standard input/output specification produce tests with empty expected outputs. We add a deterministic filter that rejects this pattern before constructing the final training corpus. The remaining errors motivate heterogeneous voting and stronger verification, but they also show that dual verification should not be interpreted as ground-truth certification.

\subsection{Task Solvability Analysis}
\label{appendix_sec:task-solvability-analysis}

\subsubsection{Solvability of Generated Problem}
\label{appendix_sec:Solvability of Generated Problem}

To estimate the fraction of potentially unsolvable problems in our generated dataset, we use GPT-5-High as a strong solver proxy. Specifically, we evaluate the pass@1 performance of several proprietary LLMs---including Qwen3-Max, Gemini-2.5-Pro, and GPT-5-High---on our voted test cases. Their single-try pass rates are reported in Table~\ref{tab:prop_llm_pass_distribution}.

\begin{table}[!h]
\centering
\scriptsize
\setlength{\tabcolsep}{3pt}
\caption{Distribution of proprietary LLMs' pass@1 on voted test cases. Each percentage range represents the fraction of tasks whose best solution from the corresponding model attains a pass rate within that interval.}
\label{tab:prop_llm_pass_distribution}
\resizebox{\columnwidth}{!}{%
\begin{tabular}{@{}lcccc@{}}
\toprule
\textbf{Range (\%)} & \textbf{R1-0528} & \textbf{Qwen3-Max} & \textbf{Gemini2.5-Pro} & \textbf{GPT5-High} \\
\midrule
(0--20)     & 13.12\% & 11.06\% & 9.57\%  & 3.07\%  \\
{[20--40)}  & 17.29\% & 16.44\% & 14.38\% & 4.83\%  \\
{[40--60)}  & 17.57\% & 18.59\% & 17.17\% & 6.49\%  \\
{[60--80)}  & 14.94\% & 16.36\% & 15.80\% & 7.80\%  \\
{[80--100)} & 13.39\% & 14.39\% & 14.90\% & 10.82\% \\
100         & 23.66\% & 23.16\% & 28.18\% & 66.98\% \\
\bottomrule
\end{tabular}%
}
\end{table}

Notably, even GPT-5-High shows a small subset of tasks with very low pass rates. Such tasks are likely to be ambiguous, underspecified, inherently unsolvable, or affected by test-case labeling noise. Since GPT-5-High is among the strongest proprietary solvers available, failures from this model serve as a practical indicator of potential flaws in the task itself.

\begin{table*}[!h]
\centering
\small
\caption{Direct contamination check between LiveCodeBench windows and training-prompt sources. \texttt{5g} and \texttt{13g} report the target-side n-gram containment score (i.e., the largest fraction of 5-token or 13-token n-grams from each LCB problem covered by any source prompt). \texttt{Emb} reports cosine similarity after MiniLM embedding re-ranking.}
\label{tab:direct_contamination}
\resizebox{0.8\textwidth}{!}{%
\begin{tabular}{@{}llccccc@{}}
\toprule
\textbf{Source} & \textbf{Window} & \textbf{\#LCB} & \textbf{\#source} & \textbf{5g p99/max} & \textbf{13g p99/max} & \textbf{Emb p99/max} \\
\midrule
X-Coder & LCB v5 & 268 & 202,128 & 0.110\,/\,0.113 & \textbf{0.030}\,/\,\textbf{0.069} & 0.826\,/\,0.871 \\
X-Coder & LCB v6 & 175 & 202,128 & \textbf{0.091}\,/\,\textbf{0.101} & 0.036\,/\,0.051 & \textbf{0.815}\,/\,\textbf{0.841} \\
OpenCodeReasoning & LCB v5 & 268 & 28,497 & 0.198\,/\,0.229 & 0.070\,/\,0.103 & 0.868\,/\,0.902 \\
OpenCodeReasoning & LCB v6 & 175 & 28,497 & 0.190\,/\,0.272 & 0.061\,/\,0.068 & 0.833\,/\,0.849 \\
\bottomrule
\end{tabular}%
}
\end{table*}

\section{Data Leakage Analysis}
\label{appendix_sec:data_leakage}

Because \method~does not directly reuse real-world tasks during post-training, its generated prompts may reduce the risk of verbatim or near-duplicate benchmark contamination relative to web-crawled task corpora. This claim concerns prompt overlap only; it does not imply independence from TACO-derived abstract features or teacher pretraining. We conduct a direct overlap analysis between the LiveCodeBench evaluation problems and the training-prompt sources used for synthetic data generation. For each LCB problem, we compute (i) \textbf{n-gram containment}: the largest fraction of its 5-token or 13-token n-grams that are covered by any single source prompt, and (ii) \textbf{embedding similarity}: cosine similarity after MiniLM embedding re-ranking over TF-IDF candidates. We examine two source corpora---X-Coder's own synthetic prompts and NVIDIA OpenCodeReasoning (which includes historical competitive-programming problems from APPS/TACO)---across both LCB v5 and v6 windows. Results are reported in Table~\ref{tab:direct_contamination}.

The X-Coder synthetic prompts exhibit consistently low lexical overlap with both LCB windows: the maximum 13-gram containment remains below 0.07, and no problem exceeds the 0.1 threshold. Embedding similarity is also moderate, with maxima below 0.88. OpenCodeReasoning serves as a stronger external baseline because it incorporates historical competitive-programming problems from APPS, TACO, and related sources; even there, overlap remains limited---only one v5 problem exceeds 0.1 in 13-gram containment, and none does so for the v6 window. Overall, these direct lexical and semantic checks confirm the absence of verbatim or near-duplicate contamination of LCB v5/v6 problems in the synthetic training prompts used by \method.

\ifcsname nolinenumbers\endcsname\nolinenumbers\fi
\section{Case Study}
\subsection{Successful Case}
\label{appendix_sec: successful case}
The SFT model frequently exhibits cognitive behaviors such as planning, verification, backtracking, and reflection, suggesting that these behaviors can be directly distilled from the teacher.
\appendixlisting[escapeinside={(*@}{@*)}]{Thinking process of successful case.}{successful_thinking.lst}

\newpage
\appendixlisting[language=Python]{Final code solution of successful case.}{successful_code.py}

\subsection{Bad Case}
\label{appendix_sec: bad case}

We identify three major failure modes in code reasoning process:

(i) Premature termination under context exhaustion.
As the context window approaches its limit, the model shortens its reasoning and rushes to produce a final answer (e.g., ``Given the time, we output the following solution..'').

(ii) Retrieval-like fallback instead of reasoning.
Rather than attempting to derive a solution, the model sometimes recalls a ``memorized" accepted submission in another language (e.g., C++) and attempts to translate it into Python, bypassing genuine reasoning.

(iii) Incomplete code emission before cutoff.
The model occasionally fails to output a complete code block before context exhaustion, leaving truncated or non-executable programs.

\begin{table*}[!t]
  \centering
  \small
  \caption{Performance analysis categorized by reasoning token length.}
  \label{tab:token_range_performance}
  \begin{tabular}{lccccc}
    \toprule
    \textbf{Token Range} & \textbf{Total} & \textbf{Passed} & \textbf{Easy} & \textbf{Medium} & \textbf{Hard} \\
    \midrule
    0--10k   & $79$ & $76$ & $46/47\ (97.9\%)$  & $22/24\ (91.7\%)$  & $8/8\ (100.0\%)$  \\
    10k--20k & $93$ & $68$ & $14/15\ (93.3\%)$  & $30/35\ (85.7\%)$  & $24/43\ (55.8\%)$ \\
    20k--32k & $96$ & $25$ & $1/1\ (100.0\%)$   & $11/27\ (40.7\%)$  & $13/68\ (19.1\%)$ \\
    \midrule
    \textbf{Total} & $\mathbf{268}$ & $\mathbf{169}$ & $\mathbf{61/63\ (96.8\%)}$ & $\mathbf{63/86\ (73.3\%)}$ & $\mathbf{45/119\ (37.8\%)}$ \\
    \bottomrule
  \end{tabular}
\end{table*}

\appendixlisting{Thinking process of bad case.}{bad_case_thinking.lst}

\subsection{Reward Hacking.}
\label{appendix_sec: reward hacking}

We manually inspect 700 RL-model rollouts from LCB v6 (175 problems, four samples each). Sixty rollouts (8.6\%) contain a hacking pattern: 56 hard-code memorized sample inputs/outputs and four emit a constant or degenerate guess.

\begin{table}[!h]
\centering
\small
\caption{Reward-hacking patterns in 700 LCB v6 rollouts.}
\label{tab:reward_hacking}
\begin{tabular}{@{}lrr@{}}
\toprule
\textbf{Pattern} & \textbf{Count} & \textbf{Share} \\
\midrule
Hard-coded sample I/O & 56 & 8.0\% \\
Constant/degenerate output & 4 & 0.6\% \\
\bottomrule
\end{tabular}
\end{table}

These behaviors rarely translate into benchmark success: 59 of the 60 hacked rollouts fail all hidden tests, while the only passing rollout also contains a correct general algorithm and the hard-coded samples are vestigial. Among 566 passing rollouts, only this one (0.18\%) contains a hack pattern. Thus hacking is visible in behavior but contributes negligibly to the reported benchmark gains.

Test-label noise can still distort training rewards. Because reward is the fraction of passed tests, a 5.27\% per-test labeling error shifts a continuous score rather than necessarily flipping an all-or-nothing label. Moreover, all eight GRPO rollouts for a prompt share the same test suite, so uniform shifts can partially cancel in relative advantages. This mechanism does not remove correlated or solution-dependent errors, which remain a limitation.

\newpage
\subsection{Extended Analysis}
\label{appendix_sec:extended_analysis}

\noindent \textbf{Cognitive Behaviors Distilled from SFT, Strategic Exploitation Emerges in RL.}
Post-SFT models exhibit emergent cognitive behaviors like planning and verification (Section~\ref{appendix_sec: successful case}), suggesting these capabilities are distilled directly from the teacher. In later RL stages, alongside performance gains, we also observe occasional instances of strategic exploitation, where the model games edge cases for partial rewards (Section~\ref{appendix_sec: reward hacking}). We also observe persistent inefficiencies, including context-induced premature termination and cross-lingual hallucinations (e.g., translating memorized C++ solutions), as detailed in Section~\ref{appendix_sec: bad case}.

\noindent \textbf{Longer Reasoning Reflects Higher Task Difficulty.}
Table~\ref{tab:token_range_performance} shows that the pass rate decreases sharply as reasoning token length increases, exhibiting a clear downward trend. This finding runs counter to the intuitive expectation that greater test-time token usage reflects deeper reasoning and should therefore yield higher accuracy.
Instead, we observe a significant chained relationship among problem difficulty, reasoning length, and pass rate: problem difficulty is positively correlated with reasoning length, while reasoning length is strongly negatively correlated with pass rate. This mediation pattern can be summarized as higher difficulty $\rightarrow$ longer reasoning length $\rightarrow$ lower pass rate.

\end{document}